\ifdefined\XeTeXversion\else\pdfoutput=1\fi
\documentclass[onecolumn]{fairmeta_contact}
\usepackage{xspace}
\usepackage{amsmath}
\usepackage{amssymb}
\usepackage{float}
\usepackage[shortlabels,inline]{enumitem}
\usepackage{currfile}
\usepackage{makecell}
\usepackage{tabularx}
\usepackage{colortbl}
\definecolor{oursblue}{HTML}{E3EFF8}
\definecolor{groupband}{HTML}{EEECF5}
\newcolumntype{Y}{>{\centering\arraybackslash}X}
\usepackage{pifont}
\newcommand{\cmark}{$\checkmark$}
\newcommand{\xmark}{$\times$}
\newcommand{\pmark}{$\circ$}

\newlength{\colw}
\titleformat*{\paragraph}{\sffamily\bfseries}
\usepackage{algorithm}
\usepackage{algpseudocode}

\title{Contact Harness: Enabling VLM Agents to Perform Contact-Rich Manipulation}
\author[*,\dagger]{Shilin Shan}
\author[*]{Yuxuan Hu}
\author[*]{Chuhao Zhou}
\author[]{Yimeng Dong}
\author[]{Bohan Hou}
\author[]{Jiaqi Bai}
\author[]{\protect\newline Boyu Ma}
\author[]{Jiahua Dong}
\author[\ddagger]{Jianfei Yang}
\affiliation[]{MARS Lab, Nanyang Technological University, Singapore}
\contribution[*]{Equal contribution}
\contribution[\dagger]{Project lead}
\contribution[\ddagger]{Corresponding author}
\metadata[Project Page]{\url{https://ntumars.github.io/ContactHarness/}}
\hypersetup{pdftitle={Contact Harness: Enabling VLM Agents to Perform Contact-Rich Manipulation},
  pdfauthor={Shilin Shan, Yuxuan Hu, Chuhao Zhou, Yimeng Dong, Bohan Hou, Jiaqi Bai, Boyu Ma, Jiahua Dong, Jianfei Yang}}

\abstract{Agents built on pretrained VLMs have demonstrated zero-shot control of real robots, but vision alone is often insufficient for reliable contact-state assessment in contact-rich manipulation. Force and tactile feedback provide complementary information, but their semantic ambiguity makes them difficult for agents to use directly. These signals must be interpreted according to their physical meaning, together with task phase and visual evidence. We propose Contact Harness, which uses reusable, optimized skills to guide a VLM agent in assessing contact states from multimodal feedback and selecting subsequent actions. A compliant execution mechanism regulates in-contact motions using real-time force feedback. The framework requires no model fine-tuning, task demonstrations, or simulation training. We conduct zero-shot real-robot evaluations using GPT-6 Astra on Rokae and Franka robots. With a single prompt skill, Contact Harness achieves a full-insertion rate of 90.7\% across 15 connector types, with further evaluations on a real PC under variations in lighting, background, and position. Across four additional contact-rich tasks, including whiteboard wiping, writing, peeling, and pivoting, Contact Harness achieves an average success rate of 100\%. On six simulated tasks, Contact Harness reaches 96.7\% success, compared with at most 68.3\% for pretrained VLA baselines fine-tuned on task demonstrations, showing the potential of agent-based manipulation for contact-rich real-world deployment.
}

\begin{document}
\addtocontents{toc}{\protect\setcounter{tocdepth}{-1}}
\maketitle
\begin{figure}[H]
  \centering
  \includegraphics[width=0.98\textwidth, trim=44 26 43 24]{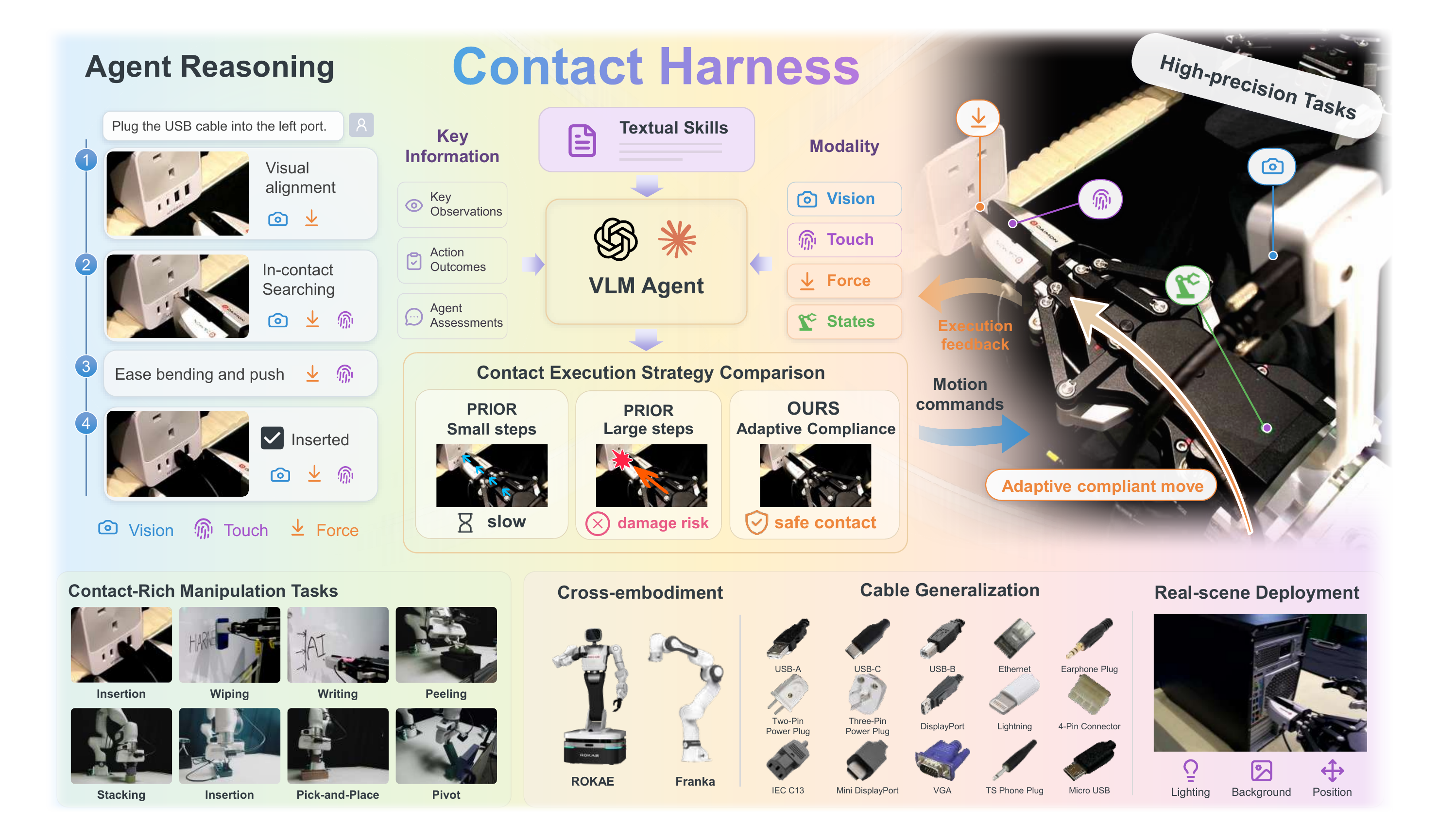}
  \caption{\textbf{Overview of Contact Harness.} A frozen VLM agent uses textual skills to assess contact states from visual, force, and tactile evidence, while compliant execution responds to contact changes between agent decisions.}
  \label{fig:teaser}
\end{figure}
\vspace{-10pt}
\section{Introduction}
\label{sec:intro}

Pretrained vision-language models (VLMs) can support robot manipulation by combining task reasoning with action generation. An agent harness connects a VLM to robot observations and control interfaces, forming a VLM agent that interacts with the physical environment. The VLM reasons over observations to determine the next action, while the harness coordinates execution and returns feedback to inform subsequent decisions~\cite{li2026k1,liu2026guava,chen2026survey}. Recent systems have demonstrated this approach across tasks such as pick-and-place and towel folding, without additional model training~\cite{jia2026agp,zhang2026astra,chen2026showharness}.

Contact-rich manipulation, however, is challenging for agents that rely primarily on visual observations~\cite{lee2019making,zhang2026astra,fu2026capx}. In cable insertion, contact feedback helps identify misalignment and regulate forces that are difficult to assess visually~\cite{spector2021insertionnet,dong2021tactile,xing2026contactdp}. Wiping~\cite{hou2025adaptive,xue2025reactive} and writing require maintaining appropriate contact force during motion. In addition, occlusion and limited visual precision make vision alone insufficient for reliable contact assessment~\cite{dong2021tactile,xing2026contactdp}.

Force and tactile feedback complement vision~\cite{he2025foar,jones2025beyond,yu2025forcevla,huang2025tactile,chen2026implicitrdp,shan2026learning}, but integrating them into VLM agents poses two challenges. First, reliable contact-state assessment requires interpreting what each signal reveals about the interaction. Increased force, for example, cannot distinguish intended contact from an unintended collision without task knowledge. Second, VLM agents typically receive execution feedback and issue new commands only after the current action or code segment finishes~\cite{chen2026showharness,zhou2026gptgear,fu2026capx}, limiting their ability to respond to contact changes during execution. Continuing a command after contact changes can produce excessive forces, particularly at high robot stiffness~\cite{will1975guarded,hogan1985impedance}.

To address these challenges, we propose \textbf{Contact Harness}, which combines multimodal contact reasoning with compliant execution to improve task and scene understanding and force regulation, thereby increasing success rates in contact-rich manipulation. Our contributions are as follows.
\begin{itemize}
\item We introduce \textbf{Contact Harness}, which uses reusable textual skills to enable a frozen VLM agent to integrate visual, force, and tactile feedback for contact-state assessment and action generation without model fine-tuning.
\item We develop a compliant execution mechanism that combines impedance control with contact-force constraints to regulate motion using real-time force feedback between successive agent decisions.
\item We evaluate Contact Harness on two real robot arms, Rokae and Franka. With a single insertion prompt, it inserts 15 connector types at a 90.7\% success rate over 150 trials, and it completes wiping, writing, peeling, and pivoting in all 40 trials. In simulation, it reaches 96.7\% success on six tasks, compared with at most 68.3\% for trained VLA policies, and ablations show that textual skills, force feedback, and compliant execution each contribute.
\end{itemize}

\section{Related work}
\label{sec:related}

Classical hybrid force-position control and impedance control provide a foundation for contact regulation in robotics~\cite{mason1981compliance,raibert1981hybrid,hogan1985impedance}. We review related work on contact policy learning, planning and program generation with large language models, and VLM-based agent harnesses.

\vspace{-10pt}

\paragraph{Contact policy learning.}
Generalist vision-language-action models learn manipulation policies from large-scale robot demonstration data~\cite{brohan2022rt,brohan2023rt, team2024octo, kim2024openvla, black2024pi0, black2025pi05, liu2025rdt, bjorck2025gr00t}. For contact-rich tasks, models acquire contact skills from interaction data~\cite{li2022seehearfeelsmart, chi2025diffusion, xue2026tube}.  Facet-0~\cite{deng2026facet0} trains a foundation model for precise manipulation using large-scale manipulation data with synchronized force measurements. ContactDP~\cite{xing2026contactdp} integrates visual, force, and tactile feedback into a diffusion policy. Some approaches combine large language model reasoning with learned control policies. A neuro-symbolic assembly method~\cite{li2026neurosymbolic} uses a large language model to generate assembly sequences and combines search with an impedance control policy trained through reinforcement learning for insertion. CompliantVLA~\cite{zhang2026compliantvla} uses a VLM to adjust stiffness and damping, while a VLA fine-tuned on task-specific data generates motion. These methods rely on learned policies.

\vspace{-10pt}

\paragraph{Program generation, planning, and control with large language models.}
Large language models can generate task plans, control programs, or motion-planning objectives from language instructions~\cite{Lin_2023,singh2022progprompt,huang2023instruct2act,rana2023sayplan,vemprala2024chatgpt,yu2023language,chen2024roboscript,huang2024rekep}. SayCan~\cite{ahn2022saycan} plans sequences of skills based on language goals and skill feasibility, while VoxPoser~\cite{huang2023voxposer} generates 3D value maps to guide motion planning. Code as Policies~\cite{liang2023cap}, CaP-X~\cite{fu2026capx}, COAP~\cite{hu2026coap}, and Act-Observe-Rewrite~\cite{kumar2026aor} investigate how large language models generate and iteratively develop robot programs. These programs can support closed-loop execution through calls to perception tools, conditional logic, and feedback loops.

For contact tasks, GenCHiP~\cite{burns2024genchip} generates compliant motion programs with termination conditions defined by force thresholds. LEMMo-Plan~\cite{chen2024lemmo} infers skill sequences and success conditions from multimodal demonstrations. Unfettered~\cite{xie2025unfettered} uses a VLM to infer desired forces, torques, and execution durations. CoRAL~\cite{cicek2026coral} uses a large language model to generate cost functions for model predictive control. These approaches translate model reasoning into skill calls, programs, parameters, or optimization objectives.

\vspace{-10pt}

\begin{figure*}[t]
\centering
\includegraphics[width=0.98\textwidth]{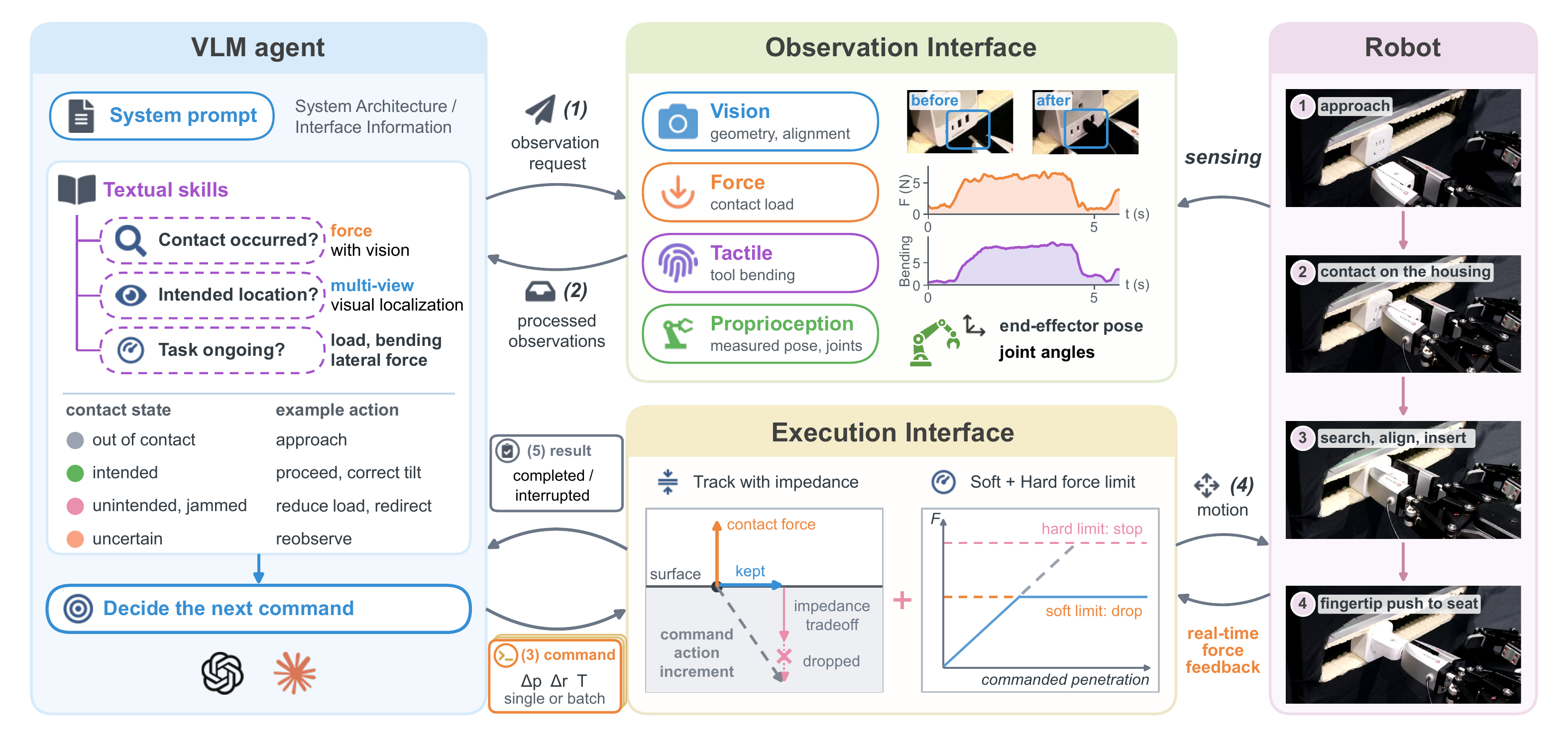}
\vspace{-8pt}
\caption{The decision and execution loop of Contact Harness. (1) The VLM requests observations, (2) assesses the contact state following the textual skills, and (3) issues a single command or a short batch. (4) During execution, real-time force feedback constrains the motion, and impedance control tracks the reference. (5) The execution result is returned to the VLM.}
\vspace{-10pt}
\label{fig:method}
\end{figure*}

\paragraph{VLM-based agent harnesses.}
An agent harness connects a model to observations and execution interfaces, coordinates command execution, and returns feedback for subsequent reasoning and action generation~\cite{yao2022react}. Some systems enable agents to generate motion commands directly or invoke primitive skills~\cite{duan2024manipulate,lima2026agentic}. Show-Harness~\cite{chen2026showharness} supports VLM-based robot control through semantic action interfaces and perception plugins. GPT-GEAR~\cite{zhou2026gptgear} evaluates GPT-6 Astra for direct closed-loop robot control, examining depth perception, in-context learning, and the use of tactile feedback for action correction. Robo-Harness K1~\cite{li2026k1} provides callable perception tools for segmentation, depth measurement, and grasp candidate generation. Other methods use language or vision-language models to detect and explain manipulation failures~\cite{liu2023reflect,duan2024aha}. Hierarchical systems combine task-level agent decisions with learned policies for action generation~\cite{shi2025hi,hu2026matters,galanti2026addressing}. Harness VLA~\cite{zhang2026harnessvla}, SafeHarness~\cite{xu2026safeharness}, and BATON~\cite{xu2026baton} use agents to coordinate task execution and invoke VLAs for manipulation, separating task-level reasoning from low-level action generation.

Beyond runtime execution, several methods use interaction experience to improve agents or manipulation policies. RPG~\cite{wang2026rpg} refines shared skills and system prompts through simulated practice. RoboFoundry~\cite{liang2026robofoundry} improves skills and context using execution experience. EmbodiRSI~\cite{lang2026embodirsi} investigates corrective data generation, while ENPIRE~\cite{xiao2026enpire} explores improvements to policy training procedures.

In contrast to these methods, our work focuses on using a frozen VLM for contact-state assessment and action generation from multimodal feedback. Contact Harness uses reusable textual skills to guide the VLM in interpreting and integrating visual, force, and tactile information, while a compliant execution mechanism regulates motion using real-time force feedback throughout each action. The supplementary material compares related methods in terms of observation, action, and contact regulation.

\section{Contact Harness}
\label{sec:method}

\subsection{Overview}
\label{sec:overview}

As shown in \cref{fig:method}, Contact Harness connects a VLM to the robot through observation and command interfaces, textual skills, and a compliant execution mechanism. Textual skills guide the VLM's interpretation of multimodal feedback and its selection of motion commands, while compliant execution uses real-time force feedback to regulate motion within each command. Each task is specified by a skill that defines its goal, essential steps, completion criteria, and task-specific decision rules.

At each iteration, the VLM requests observations, assesses the contact state, and generates a single command or a short batch of commands. Each command specifies a target displacement and an execution duration. The harness executes the commands under impedance control with contact-force constraints and returns the accepted reference poses and measured final poses. The VLM uses these results and newly requested observations to determine its next action. The interfaces and execution mechanism are shared across all tasks and robot platforms in our evaluation, while the textual skills combine a common contact-state assessment procedure with task-specific rules.

\subsection{Observation interface}

The observation interface returns the latest measurements from the sources requested by the VLM. The interface provides four complementary modalities: vision, force, tactile sensing, and proprioception. Wrist cameras and third-view cameras provide geometric cues, such as alignment error and insertion depth, but cannot directly measure contact forces. The six-axis external wrist wrench provides evidence of contact and indicates the magnitude and direction of loading, but cannot uniquely identify the contact source or location. When available, tactile sensors on both gripper fingers jointly estimate the bending load at the grasped object's tip. This signal provides an additional cue for rotational corrections, complementing the end-effector wrench estimated from joint-level signals~\cite{1570247,8059840}. Proprioception provides the end-effector pose and joint angles. Discrepancies between the measured pose and the accepted reference pose help identify motion blocked by contact.

\subsection{Textual skills for contact-state assessment}
\label{sec:text_skill}

For each task, the VLM receives a system prompt and a task skill. Although the VLM can independently reason through many operating details and task requirements, the prompts make these explicit to reduce ambiguity. This guidance is intended to reduce the reasoning effort required to determine task procedures and how to operate the system, allowing the agent to focus on selecting effective actions.

Functionally, the system prompt describes the harness architecture and the VLM's interaction with the robot. It introduces the commands for acquiring observations and executing motions, explains how to handle exceptions, and establishes operating rules shared across tasks. The task skill prompt specifies the task goal and procedure, the robot settings required for that task, and practical considerations at each stage. Together, they provide the operational context needed to use the interfaces and carry out the task.

Conceptually, these prompts guide contact-state assessment through three questions: has contact occurred, is it at the intended location, and is it proceeding as expected? The first two require interpreting force estimates alongside visual evidence, because inaccurate force estimates can mislead the assessment and force information alone cannot distinguish intended contact from unintended collisions. For the third, the agent should use force magnitude along the intended contact direction to evaluate the applied load, tactile bending to help identify tool or connector misalignment, and lateral forces in directions where free motion is expected to diagnose unintended constraints or jamming.

\subsection{Compliant execution}

A single VLM decision takes seconds, whereas the contact state can change within tens of milliseconds. Contact Harness therefore separates VLM decision making from contact regulation, using force feedback to adjust motion throughout command execution. Each command is executed through two complementary mechanisms. First, the target displacement is divided into small increments, and contact forces are assessed before each increment is applied. If the force magnitude along a robot-base axis exceeds a soft limit and the translation increment would increase loading along that axis, that component is discarded. Second, the robot tracks the resulting reference pose using impedance control~\cite{hogan1985impedance}. The reference position may lie beyond a contacted surface, with compliance allowing a position-tracking error as the robot applies load against the surface.

The VLM can adjust the soft limit according to the task phase, for example, selecting a lower value during surface search and a higher value during seating. By regulating motion within each command, this execution scheme reduces reliance on frequent VLM decisions to accommodate contact changes. For predictable free-space motions, the VLM can submit commands in batches to further reduce the number of decisions required.

\begin{figure*}[t]
\centering
\includegraphics[width=0.96\textwidth]{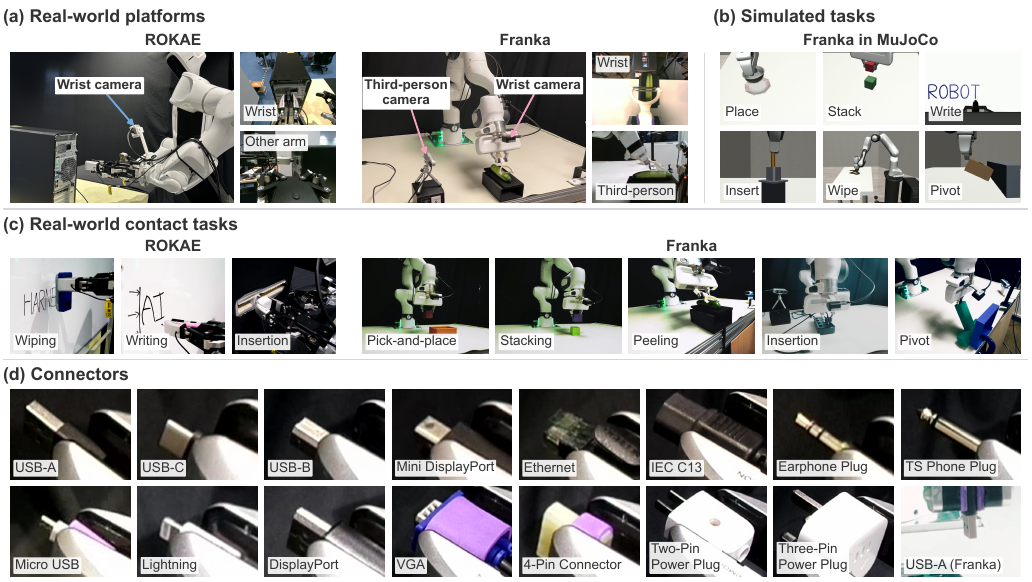}
\vspace{-5pt}
\caption{Experimental platforms and tasks. (a) The Rokae and Franka arms used in real-world experiments. (b) The six simulated tasks on a Franka arm in MuJoCo. (c) Real-world tasks on the two arms. (d) The connectors held in the Rokae gripper. The last one is inserted by the Franka arm.}
\vspace{-10pt}
\label{fig:setup}
\end{figure*}

\section{Experiments}
\label{sec:experiments}

\subsection{Setup}
\label{sec:setup}

\Cref{fig:setup} shows the robot platforms, real-world and simulation setups, and connectors. To clarify the roles of the textual skills, we distinguish between the general system prompt and task prompts throughout the remainder of this section. Their respective functions are discussed in \cref{sec:text_skill}.

\vspace{-8pt}

\paragraph{Platforms.}
We conduct real-world experiments on a 7-DoF Rokae arm and a Franka arm. The Rokae arm is equipped with a DH-Robotics AG-95 two-finger gripper. Visual observations come from three cameras: a wrist camera on the tool-holding arm, a wrist camera on the other arm providing a side view, and a head camera. Force feedback consists of the six-axis wrist wrench, transformed to the end-effector tip and then expressed in the robot-base frame for force control. Each gripper finger is equipped with a Daimon DM-Tac W2 tactile sensor. The primary tactile observation is the three-axis torque provided by the sensor API, transformed to the fingertip frame.

The Franka setup uses two cameras: a wrist camera and an external camera positioned according to task requirements. Force feedback is similarly obtained from the robot API and converted into the end-effector's frame. The Franka robot has no tactile sensors and relies on its torque estimates.

Both platforms use the same harness, observation and command interfaces, and compliant execution mechanism, with adaptations to their respective robot APIs. Simulation experiments use a Franka arm model in MuJoCo~\cite{6386109}. The same harness provides visual, force, and tactile observations, with camera placements matching the real-world setup. \cref{sec:sim} describes the simulated tasks.

\vspace{-8pt}

\paragraph{Models.}
Unless otherwise indicated, we use GPT-6 Astra with high reasoning effort in non-fast mode for real-world and simulated experiments. We also evaluate Contact Harness with Claude Fable 5.1 and 5.2 in simulation. By default, each experiment trial begins in a new session in which the agent reads only the general system prompt and a task prompt, without interaction history or additional skill files.

\begin{table*}[t]
\centering
\footnotesize
\setlength{\tabcolsep}{3pt}
\caption{\textbf{Real-robot results with GPT-6 Astra.} The upper block reports connector insertion on the Rokae arm, and the lower block reports other tasks on the Rokae and Franka arms, where PnP denotes pick-and-place. All runs use high reasoning effort, and the lower block uses the fast mode. Force is the median over trials of the peak contact force. Think.\ is the portion of time spent on model inference and observation.}
\label{tab:real}
\def\rl#1{\rotatebox{65}{\footnotesize #1}}
\def\rlref{\rl{\shortstack[l]{Mini\\DisplayPort}}}
\def\rlc#1{{\sbox0{\rlref}\sbox2{\rl{#1}}\raisebox{\dimexpr((\ht0-\dp0)-(\ht2-\dp2))/2\relax}{\usebox2}}}
\definecolor{groupband}{HTML}{EEECF5}
\definecolor{meanblue}{HTML}{E3EFF8}
\newlength\tablw\settowidth{\tablw}{Other contact}
\newcolumntype{C}{>{\hsize=0.96\hsize\centering\arraybackslash}X}\newcolumntype{W}{>{\hsize=1.6\hsize\centering\arraybackslash}X}
\begin{tabularx}{\textwidth}{p{\tablw}|*{5}{C}|*{5}{C}|*{5}{C}|W}
\toprule
\rowcolor{groupband}
\multicolumn{1}{p{\tablw}|}{\cellcolor{white}} & \multicolumn{5}{c|}{\textit{Difficulty 1}} & \multicolumn{5}{c|}{\textit{Difficulty 2}} & \multicolumn{5}{c|}{\textit{Difficulty 3}} & \multicolumn{1}{c}{} \\
\multicolumn{1}{p{\tablw}|}{{\sbox0{\rlref}\sbox2{\shortstack[l]{Connector\\insertion}}\raisebox{\dimexpr((\ht0-\dp0)-(\ht2-\dp2))/2\relax}{\usebox2}}} & \multicolumn{1}{c}{\hspace*{3pt}\makebox[0pt][c]{\rlc{\shortstack[l]{TS Phone\\Plug}}}\hspace*{-3pt}} & \multicolumn{1}{c}{\makebox[0pt][c]{\rlc{\shortstack[l]{Three-Pin\\Power Plug}}}} & \multicolumn{1}{c}{\makebox[0pt][c]{\rlc{\shortstack[l]{Two-Pin\\Power Plug}}}} & \multicolumn{1}{c}{\makebox[0pt][c]{\rlc{IEC C13}}} & \multicolumn{1}{c|}{\makebox[0pt][c]{\rlc{USB-A}}} & \multicolumn{1}{c}{\makebox[0pt][c]{\rlc{Ethernet}}} & \multicolumn{1}{c}{\makebox[0pt][c]{\rlc{USB-C}}} & \multicolumn{1}{c}{\makebox[0pt][c]{\rlc{\shortstack[l]{Earphone\\Plug}}}} & \multicolumn{1}{c}{\makebox[0pt][c]{\rlc{Lightning}}} & \multicolumn{1}{c|}{\hspace*{-4pt}\makebox[0pt][c]{\rlc{\shortstack[l]{4-Pin\\Connector}}}\hspace*{4pt}} & \multicolumn{1}{c}{\makebox[0pt][c]{\rlc{USB-B}}} & \multicolumn{1}{c}{\makebox[0pt][c]{\rlc{VGA}}} & \multicolumn{1}{c}{\makebox[0pt][c]{\rlc{\shortstack[l]{Mini\\DisplayPort}}}} & \multicolumn{1}{c}{\makebox[0pt][c]{\rlc{DisplayPort}}} & \multicolumn{1}{c|}{\makebox[0pt][c]{\rlc{\shortstack[l]{Micro\\USB}}}} & \multicolumn{1}{c}{\sbox0{\rl{\shortstack[l]{Three-Pin\\Power Plug}}}\raisebox{\dimexpr(\ht0-\height)/2+0.55\baselineskip\relax}{Mean}} \\
\midrule
\rowcolor{oursblue}
Success & 10/10 & 10/10 & 10/10 & 10/10 & 10/10 & 9/10 & 10/10 & 10/10 & 9/10 & 9/10 & 8/10 & 9/10 & 7/10 & 8/10 & 7/10 & \textbf{136/150} \\
\specialrule{\cmidrulewidth}{0pt}{0pt}
\rule{0pt}{2.5ex}Time (s) & 291 & 336 & 335 & 315 & 365 & 273 & 366 & 489 & 502 & 408 & 474 & 313 & 301 & 417 & 482 & 378 \\
Decisions & 19.2 & 18.5 & 19.4 & 19.3 & 20.4 & 17.7 & 25.1 & 23.8 & 24.5 & 25.7 & 29.3 & 25.3 & 22.6 & 24.9 & 30.1 & 23.1 \\
Force (N) & 15.7 & 22.5 & 23.9 & 23.5 & 18.4 & 15.0 & 17.6 & 18.6 & 18.8 & 19.9 & 23.2 & 24.0 & 20.2 & 20.4 & 17.6 & 20.0 \\
Think. (\%) & 89.9 & 91.6 & 91.5 & 90.5 & 92.1 & 91.1 & 89.7 & 91.6 & 92.4 & 91.1 & 91.3 & 89.6 & 83.4 & 91.3 & 88.4 & 90.4 \\
\midrule
\end{tabularx}\par\vspace{-\arrayrulewidth}
\newlength\tabX\setlength\tabX{\dimexpr(\textwidth-\tablw-34\tabcolsep-4\arrayrulewidth)/16\relax}
\newlength\tabXX\setlength\tabXX{\dimexpr2\tabX+2\tabcolsep\relax}
\begin{tabularx}{\textwidth}{p{\tablw}|*{2}{Y}|Y|*{5}{Y}|Y|Y}
\rowcolor{groupband}
\multicolumn{1}{p{\tablw}|}{\cellcolor{white}} & \multicolumn{3}{c|}{\textit{Rokae}} & \multicolumn{6}{c|}{\textit{Franka}} & \multicolumn{1}{c}{} \\
\multirow{-2}{*}{\shortstack[l]{Other\\tasks}} & \multicolumn{1}{c}{Wiping} & \multicolumn{1}{c}{Writing} & \multicolumn{1}{c|}{Mean} & \multicolumn{1}{c}{Insertion} & \multicolumn{1}{c}{Peeling} & \multicolumn{1}{c}{Pivoting} & \multicolumn{1}{c}{Stacking} & \multicolumn{1}{c}{PnP} & \multicolumn{1}{c|}{Mean} & \multicolumn{1}{c}{Mean} \\
\midrule
\rowcolor{oursblue}
Success & 10/10 & 10/10 & \textbf{20/20} & 10/10 & 10/10 & 10/10 & 10/10 & 10/10 & \textbf{50/50} & \textbf{70/70} \\
\specialrule{\cmidrulewidth}{0pt}{0pt}
\rule{0pt}{2.5ex}Time (s) & 308 & 413 & 360 & 203 & 227 & 266 & 284 & 357 & 267 & 294 \\
Decisions & 25.5 & 42.6 & 34.1 & 15.1 & 13.6 & 30.0 & 15.3 & 18.1 & 18.4 & 22.9 \\
Force (N) & 15.5 & 10.8 & 13.2 & 18.4 & 14.2 & 17.1 & 11.2 & 7.5 & 13.7 & 13.5 \\
Think. (\%) & 56.2 & 70.5 & 63.4 & 82.5 & 82.0 & 70.9 & 75.3 & 75.1 & 77.2 & 73.2 \\
\bottomrule
\end{tabularx}
\end{table*}

\begin{table}[t]
\begin{minipage}[t]{\colw}
\centering
\small
\setlength{\tabcolsep}{2.5pt}
\caption{\textbf{Insertion on a real PC} with GPT-6 Astra (high, fast mode) on the Rokae robot, with lighting, background, and socket position changed together. Each connector has 10 trials. With memory, the marked trials run in one agent conversation.}
\label{tab:realpc}
\begin{tabular*}{\linewidth}{@{\extracolsep{\fill}}lcccc}
\toprule
& IEC & Earphone Plug & USB-A & Mean \\
\midrule
Memory    & last 5 trials & last 5 trials & \checkmark & / \\
Success   & 10/10 & 9/10 & 10/10 & 29/30 \\
Time (s)  & 377 & 269 & 200 & 282 \\
Decisions & 26.9 & 19.4 & 19.0 & 21.8 \\
Force (N) & 25.4 & 20.7 & 24.5 & 23.5 \\
Think. (\%) & 87.2 & 86.1 & 82.5 & 85.3 \\
\bottomrule
\end{tabular*}
\end{minipage}\hfill
\begin{minipage}[t]{\colw}
\centering
\small
\setlength{\tabcolsep}{3pt}
\caption{\textbf{Speed tests} with GPT-6 Astra (high). IEC E1 is copied from \cref{tab:real}. IEC E2 and E3 are evaluated on the Rokae arm, and USB-A on the Franka arm. Fast denotes the fast mode. With memory, the 10 trials run in one agent conversation.}
\label{tab:speed}
\begin{tabular*}{\linewidth}{@{\extracolsep{\fill}}lcccc}
\toprule
Connector & IEC E1 & IEC E2 & IEC E3 & USB-A \\
\midrule
Fast      & \xmark & \checkmark & \checkmark & \checkmark \\
Memory    & \xmark & \xmark & \checkmark & \checkmark \\
Success   & 10/10 & 10/10      & 10/10      & 10/10 \\
Time (s)  & 315   & 183        & 96         & 147 \\
Decisions & 19.3  & 17.9       & 10.4       & 12.3 \\
Think. (\%) & 90.5 & 85.8      & 83.1       & 79.7 \\
\bottomrule
\end{tabular*}
\end{minipage}
\end{table}

\vspace{-8pt}

\paragraph{Protocol and metrics.}
(1) Insertion: Each connector insertion trial begins with the connector housing held in the gripper. The arm starts from a pose in front of the target, with the sockets visible in the wrist-camera view. An insertion is considered successful only when the connector head reaches its mechanical insertion limit within the socket; apparent seating in the camera view alone is insufficient. Each trial has a 15-minute time limit.
(2) Whiteboard wiping: Each trial begins with the eraser held in the gripper. Success requires removing all marks. Simulated wiping uses the same criterion.
(3) Writing: Each trial begins with the marker pen held in the gripper. Success requires the written word to be recognizable. Simulated writing uses the same criterion.
(4) Vegetable peeling: Each trial begins with the peeler held in the gripper and secured with additional supports to compensate for limited friction at the gripper-peeler interface and prevent slipping during contact. Success requires peeling the segment specified at the beginning of the conversation or, if no target segment is specified, removing a strip of skin with uniform width and thickness.
(5) Pivoting: Success requires the box or T-shaped object to be pivoted and retain the resulting orientation after the gripper withdraws.
(6) Stacking and pick-and-place: Success requires the grasped cube to remain on the base cube or in the basket after the gripper releases.

We record two efficiency metrics for each trial. Execution time is measured from task receipt until the agent reports completion. The number of decisions counts VLM or policy calls that generate actions, with either a single command or a batch of commands counted as one decision. This count helps assess reasoning efficiency when fluctuations in Astra and Fable service latency introduce substantial delays. Both metrics are averaged over successful trials. Each experimental condition is repeated 10 times.

\subsection{Real-robot evaluation}
\label{sec:real}

\paragraph{Prompt usage.}

The system prompt is shared across all tasks on a given robot. Each task uses the same task prompt across robot platforms, with minor adjustments to describe the physical setup and robot configuration, such as camera orientations and observation sources. Detailed prompts for real-world and simulated tasks are provided in the appendix.

\vspace{-5pt}

\paragraph{Insertion across 15 connectors.}

We categorize the connectors into three difficulty levels based on the following considerations. (1) Easy: the connectors are sufficiently large in the camera view, have simple alignment requirements, and are either short or relatively loose-fitting, making jamming less likely. (2) Medium: the connectors have sharp corners or challenging geometries, or appear small in the camera view, requiring repeated alignment adjustments. (3) Hard: the connectors are either very tight-fitting or visually very small, requiring repeated searching and precise axial alignment during insertion to avoid jamming.

The upper block of \cref{tab:real} reports insertion results for 15 connector types on the Rokae arm using GPT-6 Astra with high reasoning effort, with 10 trials per connector. Full-insertion success rates are 100\%, 94\%, and 78\% for the easy, medium, and hard categories, respectively, averaging 90.7\% across all 15 connector types. Wall-clock execution time is strongly affected by fluctuations in Astra and Fable service latency. The decision count reflects the effort required to resolve ambiguities in the physical scene and the effectiveness of the resulting motion commands.

\vspace{-5pt}

\paragraph{Additional contact tasks.}
The lower block of \cref{tab:real} reports results on additional tasks using GPT-6 Astra with high reasoning effort, with 10 trials per task. The Rokae arm is used for whiteboard wiping and writing, and the Franka arm for peeling, pivoting, and insertion cross-validation. We also evaluate pick-and-place and cube stacking on Franka to test Contact Harness on contact-free tasks.

Each task uses a task-specific prompt that is consistent across robot platforms. Success criteria are defined in \cref{sec:setup}. These tasks impose different contact requirements. Wiping and writing require maintaining a stable contact force while moving along a surface. Peeling requires maintaining a shallow cutting depth along a curved surface. Pivoting requires maintaining a stable contact-force magnitude as the force direction changes over time. The average success rate is 100\% across the tasks on Rokae and 100\% across the three contact tasks on Franka.

\vspace{-5pt}

\begin{table*}[t]
\centering
\small
\setlength{\tabcolsep}{3pt}
\caption{\textbf{Results on the six simulated tasks.} VLM-based baselines use GPT-6 Astra. Units: Succ.\ in \%, time in s, force in N. Time and decisions are means over successful runs; $^*$excluding Write. Force is the median over runs of the peak contact force sustained for 0.2\,s, which excludes brief impact spikes. The four right-hand columns are averaged over tasks. w.cam: with the wrist camera. goal only: only the task goal sentence. our prompt skill: our task prompts, additionally adapted to each method.}
\label{tab:sim}
\begin{tabularx}{\textwidth}{ll|*{6}{Y}|*{4}{Y}}
\toprule
\rowcolor{groupband}
\multicolumn{2}{l|}{\cellcolor{white}} & \multicolumn{6}{c|}{\textit{Success}} & \multicolumn{4}{c}{\textit{Mean over tasks}} \\
& \multirow{-2}{*}{Method} & Place & Stack & Insert & Wipe & Write & Pivot & Succ.$\uparrow$ & Time$^*$$\downarrow$ & Dec.$^*$ & Force$\downarrow$ \\
\midrule
 & CaP-X~\cite{fu2026capx} & 5/10 & 3/10 & 1/10 & 4/10 & 1/10 & 0/10 & 23.3 & 669 & 8.4 & 51.2 \\
 & CaP-X + w.cam & 2/10 & 1/10 & 0/10 & 6/10 & 1/10 & 0/10 & 16.7 & 489 & 7.9 & 137.2 \\
 & CaP-X + our prompt skill & 0/10 & 0/10 & 0/10 & 0/10 & 0/10 & 0/10 & 0.0 & -- & -- & 28.4 \\
 & GenCHiP (goal only) & 9/10 & 8/10 & 0/10 & 1/10 & 2/10 & 0/10 & 33.3 & 118 & 1.0 & 11.0 \\
\multirow{-5}{*}{\makecell[l]{\textit{Code as}\\\textit{policies}}} & GenCHiP + our prompt skill & 5/10 & 5/10 & 0/10 & 0/10 & 0/10 & 0/10 & 16.7 & 94 & 1.0 & 7.3 \\
\midrule
 & $\pi_{0.5}$~\cite{black2025pi05} & 5/10 & 6/10 & 0/10 & 3/10 & 5/10 & 9/10 &  46.7 & 37 & 62.9 & 7.8   \\
 & $\pi_{0.5}$ + force & 8/10 & 6/10 & 1/10 & 4/10 & 8/10 & 10/10 &  61.7 & 30 & 47.4 & 10.0   \\
 & ForceVLA~\cite{yu2025forcevla} & 10/10 & 8/10 & 2/10 & 6/10 & 6/10 & 9/10 &  68.3 & 49 & 53.6 & 11.3   \\
\multirow{-4}{*}{\makecell[l]{\textit{Training-}\\\textit{based}\\\textit{policies}}} & CompliantVLA~\cite{zhang2026compliantvla} & 9/10 & 6/10 & 0/10 & 3/10 & 0/10 & 8/10 &  43.3 & 240 & 45.3 & 9.6   \\
\midrule
\textit{Harness} & Show-Harness & 9/10 & 1/10 & 0/10 & 0/10 & 0/10 & 0/10 & 16.7 & 906 & 57.3 & 9.6 \\
\midrule
\rowcolor{oursblue}
& Contact Harness (Astra) & 10/10 & 10/10 & 8/10 & 10/10 & 10/10 & 10/10 & \textbf{96.7} & 331 & 26.3 & 7.3 \\
& Contact Harness (Fable 5.1) & 10/10 & 10/10 & 9/10 & 10/10 & 10/10 & 10/10 & \textbf{98.3} & 436 & 15.0 & 7.8 \\
\multirow{-3}{*}{\textbf{\textit{Ours}}} & Contact Harness (Fable 5.2) & 10/10 & 10/10 & 9/10 & 10/10 & 10/10 & 10/10 &  \textbf{98.3} & 432 & 13.8 & 7.6  \\
\bottomrule
\end{tabularx}
\end{table*}

\paragraph{Zero-shot generalization and execution speed potential.}
In the preceding experiments, we choose the positions of sockets and other interaction targets without deliberate optimization for performance, relying on the agent's zero-shot generalization to adapt. To further assess the potential for real-world deployment, we apply Contact Harness directly to connector insertion on a real PC, simultaneously changing the lighting, background, and socket positions. We test three connectors with 10 trials each. As shown in \cref{tab:realpc}, the agent succeeds in 29 of 30 trials.

We further conduct three preliminary speed tests using GPT-6 Astra in fast mode with high reasoning effort to assess the speed potential of Contact Harness under relatively stable service conditions with prioritized processing. In two of the three tests, the agent performs 10 consecutive insertions within a single conversation, with manual resets between insertions. This allows it to reuse previous motion patterns instead of searching for the socket from scratch each time. As reported in \cref{tab:speed}, all 30 trials succeed, with average completion times per trial of 96 seconds for IEC plug insertion and 147 seconds for USB-A insertion. This setting more closely reflects real-world deployment, where interaction history can guide subsequent operations.

\subsection{Simulated benchmark}
\label{sec:sim}

\paragraph{Tasks.}
The simulation benchmark comprises four contact tasks and two contact-free tasks, shown in \cref{fig:setup}. (1) Place and Stack serve as contact-free pick-and-place reference tasks. Place requires placing a bowl on a plate, while Stack uses the robosuite~\cite{zhu2020robosuite} Stack task, in which the robot stacks a red cube on a green cube. (2) Insert is modified from the Isaac Lab~\cite{mittal2025isaac} Factory~\cite{narang2022factory} PegInsert task and rebuilt in MuJoCo, with details in the supplementary material. We make the task stricter by adding press-fit resistance: beyond an insertion depth of 3\,mm, the peg advances only under a sustained axial force above 15\,N, reflecting the force requirements of real connectors. As in the original task, an episode succeeds only when the peg is within 1\,mm of the hole bottom. (3) Wipe uses the robosuite Wipe task, in which a mounted wiping pad must remove all 100 dirt markers. (4) Write is adapted from the ManiSkill3~\cite{tao2024maniskill3} DrawSVG task and requires writing `ROBOT' on a whiteboard with a grasped marker, with success judged by visual inspection. (5) Pivot is a new task in which the robot pushes a box using the gripper against a low wall to rotate the box.

For comparisons with other methods, success is determined solely by the task conditions above, without an additional force-based failure criterion. In the ablation experiments presented later, however, a run also fails if any specified force constraint is violated, for example, if a contact force above 50\,N is sustained for at least 0.2\,s. Detailed criteria are provided in the appendix.

\vspace{-8pt}

\paragraph{Comparison.}

We compare Contact Harness with agentic methods, including CaP-X~\cite{fu2026capx}, GenCHiP~\cite{burns2024genchip}, and Show-Harness~\cite{chen2026showharness}, under different settings. We also collect dedicated data for the free-space and contact-rich tasks evaluated in this paper and fine-tune VLA-based approaches, including $\pi_{0.5}$~\cite{black2025pi05}, ForceVLA~\cite{yu2025forcevla}, and CompliantVLA~\cite{zhang2026compliantvla}. Quantitative results are reported in \cref{tab:sim}.

On the four contact-rich tasks, CaP-X and GenCHiP achieve low overall success rates of 15.0\% and 7.5\%, respectively. A common failure is that the generated trajectories lack the resolution and compliance needed to maintain contact forces while making the required positional adjustments. Notably, adding task descriptions extracted from our textual skills does not improve their performance. This results from the additional caution induced by contact-related prompts, which makes the agent more conservative when generating code, slows task progress, and leads to failures when the limit on code generation runs is reached.

Show-Harness was originally designed for pick-and-place tasks and uses a discretized action space with a minimum translational step of 1 cm, which may limit its applicability to contact-rich manipulation. Despite this discretization, we evaluate it under impedance control to assess whether compliance can support in-contact motion. It achieves an overall success rate of 16.7\%, with higher success on pick-and-place and stacking, suggesting that coarse action discretization is indeed a limitation for contact-rich tasks. The appendix provides implementation details, our adaptations of these methods to contact-rich tasks, and detailed failure analyses.

For policy-based action generation, we evaluate $\pi_{0.5}$ with different observation modalities, alongside ForceVLA and CompliantVLA. We collect 50 trials per task in MuJoCo to train these models, with detailed training parameters provided in the appendix. The training-based approaches perform better overall, with ForceVLA achieving a success rate of 68.3\%. However, even with expert demonstrations, insertion success is still low under the deliberately challenging simulated insertion conditions.

Contact Harness achieves a success rate above 96\% across three candidate agent models. We attribute this performance primarily to the combination of impedance control and additional contact-force regulation, flexible action lengths guided by the textual skills, and frequent observations and contact-state assessments. Together, these components enable timely responses and adjustments to unexpected changes in contact.

\vspace{-8pt}

\begin{table*}[t]
\centering
\small
\setlength{\tabcolsep}{3pt}
\caption{\textbf{Ablation on the four simulated contact tasks} with GPT-6 Astra, ten runs per task. Units: Succ.\ and Think.\ in \%, time in s, force in N. Each of the ablation row removes one component. Without compliant execution and the soft force limit, motions run under stiff position control with no force limit. $^\dagger$The wrench reading is hidden only from the agent, while compliant execution and the force limits still use it. $^\ddagger$Successful insertions arise from different phases, as discussed in the ablation paragraph.}
\label{tab:ablation}
\begin{tabularx}{\textwidth}{l|*{4}{Y}|*{5}{Y}}
\toprule
\rowcolor{groupband}
\multicolumn{1}{l|}{\cellcolor{white}} & \multicolumn{4}{c|}{\textit{Success}} & \multicolumn{5}{c}{\textit{Mean over tasks}} \\
\multirow{-2}{*}{Configuration} & Insert & Wipe & Write & Pivot & Succ.$\uparrow$ & Time$\downarrow$ & Dec. & Force$\downarrow$ & Think. \\
\midrule
\rowcolor{oursblue}
Contact Harness (full) & 8/10$^\ddagger$ & 10/10 & 10/10 & 10/10 & \textbf{95.0} & 367 & 28.2 & 9.7 & 79.7  \\
w/o force observations$^\dagger$ & 7/10$^\ddagger$ & 7/10 & 9/10 & 10/10 &  82.5 & 367 & 26.5 & 8.5 & 78.1   \\
w/o skills & 0/10 & 0/10 & 9/10 & 8/10 &  42.5 & 221 & 12.3 & 5.6 & 68.2   \\
w/o compliance \& soft limit & 0/10 & 9/10 & 10/10 & 6/10 &  62.5 & 279 & 27.7 & 28.6 & 81.0   \\
\bottomrule
\end{tabularx}
\end{table*}

\paragraph{Ablation.}

We evaluate the contributions of individual components by separately removing force observations from the agent's inputs, textual skill injection, and low-level compliance and force constraints. Results are reported in \cref{tab:ablation}.

Removing the textual skills causes Insert and Wipe to fail completely. Without low-level compliance, Wipe still succeeds, but the average peak force increases to nearly 30\,N, compared with 9.7\,N for the full method. When force observations are withheld from the agent, success rates are comparable. However, successful insertions under this condition arise through entirely different mechanisms from those in the full method. Seven of the eight successful trials with the full method achieve insertion during the alignment phase, in which the agent is instructed to use wrench observations to reduce undesirable forces and torques while advancing the peg. In contrast, all seven successful trials without force observations fail to complete during this phase and instead rely on the forced final push specified in the task prompt. This suggests that, despite compliance, visual evidence alone is still insufficient for the agent to make decisions and perform force-guided corrections without wrench information.

We further examine the agents' reasoning effort. Across \cref{tab:real,tab:realpc,tab:speed}, reasoning accounts for the majority of task execution time, averaging 84.7\% across all experiments. These results suggest that minimizing reasoning latency, for example to the millisecond scale, while preserving the agent's generalization capability could allow agentic approaches to complete tasks in approximately $1/6$ of the several minutes currently required, while maintaining high success rates.

\section{Conclusion and Limitations}
\label{sec:conclusion}

We presented Contact Harness, which enables a pretrained VLM agent to perform contact-rich manipulation by combining multimodal contact assessment with compliant execution. The framework requires no additional training or fine-tuning. Experiments on Rokae and Franka robots demonstrated its capabilities in connector insertion, wiping, writing, peeling, and pivoting. Six simulated tasks provided baseline comparisons and evaluations across VLMs, with Contact Harness showing high success rates.

\vspace{-8pt}
\paragraph{Limitations.}\label{sec:limitations} The main limitation is long execution time, primarily due to the agent's reasoning latency. Although average insertion time can be reduced to 96 seconds (\cref{tab:speed}), execution remains slow compared with policy-based approaches. Additional challenges in deploying Contact Harness in complex real-world settings are discussed in the supplementary material.

\section*{Acknowledgments}
This work is jointly supported by MOE Singapore Tier 1 Grant and a Start-up Grant from Nanyang Technological University. We thank ROKAE Robotics for technical support with the robot setup and the real-robot experiments.

\section*{LLM Usage Statement}
The authors take full responsibility for all submitted content, regardless of any LLM or AI-tool use. LLM-based tools (Claude and ChatGPT) were used only at the assistance level, that is, for editing or refining author-produced content, in four parts of this work: writing and editing the text, code and implementation, the experiments (design, execution, and analysis of results), and figures and other visual material. No LLM or AI tool was used for related work and citations, for data (synthetic data, dataset creation, or annotation), or for research ideas, hypotheses, claims, or conclusions. Neither the full manuscript nor major parts of it were given to an LLM provider to obtain reviewer-style feedback. The VLMs that act as the agent in Contact Harness and in the baselines are part of the studied methods and are described in the main text.

\bibliographystyle{ieeenat_fullname}
\bibliography{main}
\clearpage
\beginappendix
\addtocontents{toc}{\protect\setcounter{tocdepth}{2}}
\setcounter{figure}{0}\setcounter{table}{0}
\renewcommand{\theHfigure}{S\arabic{figure}}
\renewcommand{\theHtable}{S\arabic{table}}
\renewcommand{\thesection}{\Alph{section}}
\renewcommand{\thetable}{S\arabic{table}}
\renewcommand{\thefigure}{S\arabic{figure}}
\setcounter{tocdepth}{2}
\raggedbottom

\makeatletter
\@starttoc{toc}
\makeatother

\vspace{0.6em}
The supplementary material is organized as follows. \Cref{sec:s_rw} provides a point-by-point comparison of Contact Harness with related work. \Cref{sec:s_sys} describes the hardware, compliant execution mechanism, observation interface, and command interface for the ROKAE and Franka platforms. \Cref{sec:s_skill} explains the composition of the textual skills using an example. \Cref{sec:s_real} provides the failure analysis, the recorded observation requests in each phase, and execution sequences for the real-robot experiments. \Cref{sec:s_sim} covers the simulation benchmark, including task construction, success criteria, baseline implementations, the benchmark with its VLA baselines, Contact Harness results, baseline failure analyses, and ablation experiments. \Cref{sec:s_example} presents complete execution records for one simulated run and one real-robot run. \Cref{sec:s_token} reports token usage on the real robot and in simulation. \Cref{sec:s_limit} discusses limitations, and \cref{sec:s_prompts} summarizes the prompts for all tasks.

\section{Comparison with Related Work}
\label{sec:s_rw}
\begin{table*}[tp]
\centering
\footnotesize
\setlength{\tabcolsep}{3pt}
\caption{Comparison with related work. \pmark{} means partially supported, and \xmark{} means not supported or not stated in the paper.}
\label{tab:related}
\renewcommand{\arraystretch}{1.28}%
\renewcommand{\tabularxcolumn}[1]{m{#1}}
\begin{tabularx}{\textwidth}{m{1.95cm}m{2.95cm}|*{4}{>{\centering\arraybackslash}m{1.4cm}}|X}
\toprule
\rowcolor{groupband}
\parbox[c][4.4\normalbaselineskip][c]{1.95cm}{Category} & \parbox[c][4.4\normalbaselineskip][c]{2.95cm}{Method} & \parbox[c][4.4\normalbaselineskip][c]{1.4cm}{\centering Stepwise\\LLM/VLM\\decisions\\in contact} & \parbox[c][4.4\normalbaselineskip][c]{1.4cm}{\centering Force or\\tactile in\\decisions} & \parbox[c][4.4\normalbaselineskip][c]{1.4cm}{\centering Compliant\\execution} & \parbox[c][4.4\normalbaselineskip][c]{1.4cm}{\centering Training-\\free and\\demo-free} & \parbox[c][4.4\normalbaselineskip][c]{3cm}{Real-robot contact tasks} \\
\midrule
 & $\pi_{0.5}$~\cite{black2025pi05} & \xmark & \xmark & \xmark & \xmark & Household tasks such as wiping spills and closing cabinets \\
 & ForceVLA~\cite{yu2025forcevla} & \xmark & \cmark & \xmark & \xmark & Plug and USB drive insertion, bottle pumping, whiteboard wiping, cucumber peeling \\
 & Facet-0~\cite{deng2026facet0} & \xmark & \cmark & \cmark & \xmark & PC component assembly (GPU, RAM, drive, CPU) \\
 & ContactDP~\cite{xing2026contactdp} & \xmark & \cmark & \cmark & \xmark & Industrial connector insertion \\
 & \makecell[l]{Neuro-symbolic\\assembly~\cite{li2026neurosymbolic}} & \xmark & \cmark & \cmark & \xmark & {Multi-pair peg-in-hole} and gear assembly \\
\multirow{-6}{*}{\makecell[l]{\textit{Trained}\\\textit{networks}}} & CompliantVLA~\cite{zhang2026compliantvla} & {\pmark} & \cmark & \cmark & \xmark & Box pushing, whiteboard wiping \\
\midrule
 & CaP~\cite{liang2023cap} & \xmark & \xmark & \xmark & \cmark & Drawing and erasing shapes on a whiteboard (qualitative only) \\
 & CaP-X~\cite{fu2026capx} & \pmark & {\pmark} & \xmark & \cmark & {Elevator button pressing (qualitative only)} \\
 & COAP~\cite{hu2026coap} & \xmark & \xmark & \xmark & \cmark & None (simulation only) \\
 & GenCHiP~\cite{burns2024genchip} & \xmark & Threshold & \cmark & \cmark & Peg insertion, connector insertion, cable routing \\
 & LEMMo-Plan~\cite{chen2024lemmo} & \pmark & \makecell{Threshold\\(demos)} & \cmark & \xmark & {Cable mounting, cap tightening} \\
 & Unfettered~\cite{xie2025unfettered} & \pmark & \makecell{Force\\commands} & \cmark & \cmark & Pushing, opening and closing lids \\
\multirow{-7}{*}[1.0\normalbaselineskip]{\makecell[l]{\textit{LLM-generated}\\\textit{programs,}\\\textit{plans, or}\\\textit{objectives}}} & CoRAL~\cite{cicek2026coral} & \pmark & \cmark & \cmark & \cmark & Pushing, flipping, pick-and-place \\
\midrule
 & Harness VLA~\cite{zhang2026harnessvla} & \pmark & \xmark & \xmark & \xmark & Wiping a plate with a sponge, pouring steel {beads} (qualitative only) \\
 & SafeHarness~\cite{xu2026safeharness} & \pmark & \xmark & \xmark & \xmark & None (pick-and-place) \\
 & BATON~\cite{xu2026baton} & \pmark & \xmark & \xmark & \xmark & None (pick-and-place) \\
 & RoboFoundry~\cite{liang2026robofoundry} & {\pmark} & \xmark & \xmark & \xmark & Nesting doll insertion (executed by a fine-tuned $\pi_{0.5}$) \\
 & EmbodiRSI~\cite{lang2026embodirsi} & \xmark & \xmark & \xmark & \xmark & None (pick-and-place) \\
\multirow{-6}{*}{\makecell[l]{\textit{Agent harness}\\\textit{with actions}\\\textit{from trained}\\\textit{policies}}} & ENPIRE~\cite{xiao2026enpire} & \xmark & \xmark & \xmark & \xmark & Pin insertion, GPU installation, push-T, zip tie cutting \\
\midrule
 & AGP~\cite{jia2026agp} & \cmark & \xmark & \xmark & \pmark & Part assembly (including pin insertion) \\
 & GPT-6 Astra~\cite{zhang2026astra} & \cmark & \xmark & \xmark & \cmark & Tube insertion, charger insertion (0/3 each) \\
 & GPT-GEAR~\cite{zhou2026gptgear} & \cmark & \makecell{\cmark\\(simulation)} & \xmark & \cmark & {Block} insertion into a cup, pouring (2/5 each, zero-shot) \\
 & Show-Harness~\cite{chen2026showharness} & \cmark & \xmark & \pmark & \cmark & Peg insertion (also pick-and-place and stacking) \\
 & Robo-Harness K1~\cite{li2026k1} & \cmark & \xmark & \xmark & \cmark & None (simulation only) \\
\multirow{-6}{*}[0.5\normalbaselineskip]{\makecell[l]{\textit{Agent harness}\\\textit{generating}\\\textit{actions directly}}} & RPG~\cite{wang2026rpg} & \pmark & \xmark & \xmark & \pmark & Cylinder insertion, whiteboard wiping (qualitative only) \\
\midrule
\rowcolor{oursblue}
\textbf{\textit{Ours}} & \textbf{Contact Harness} & \cmark & \cmark & \cmark & \cmark & \textbf{Insertion of 15 connector types, whiteboard wiping, writing, peeling, pivoting} \\
\bottomrule
\end{tabularx}
\end{table*}

\Cref{tab:related} lists representative works from the three categories in Sec.~2 of the main paper and compares them along four dimensions. The comparison is based on each paper's real-robot system, with simulation results explicitly marked in the table. \emph{Stepwise LLM/VLM decisions in contact} means that, after contact occurs, a general-purpose LLM or VLM uses the latest observations to decide the next action. \emph{Force or tactile in decisions} means that the model selecting actions has access to force or tactile information as an input or observation modality. This column includes two additional annotations: \emph{Threshold} indicates that the model-generated program uses force thresholds as termination conditions, while \emph{force commands} indicates that the model outputs desired forces for a controller to track. \emph{Compliant execution} means that actions are executed under compliant control. \emph{Training-free and demo-free} means that the method uses no robot data for training and requires no task demonstrations. The final column lists the contact tasks evaluated on real robots in each work.

The methods in the table exhibit two complementary emphases. In one group, an LLM or VLM selects each action during contact based on the latest observations. On real robots these observations are visual, and force or tactile feedback does not enter the decisions. The other group uses compliant execution to regulate contact forces, while actions typically come from a trained policy or a program generated in advance, without the LLM or VLM directly selecting each action during contact. The first group is strong in decision making, while the second is strong in execution. Contact Harness combines these strengths by pairing compliant execution with stepwise VLM decisions informed by visual, force, and tactile observations.

\section{System Implementation on the Rokae and Franka Platforms}
\label{sec:s_sys}
Both platforms use the same Contact Harness. \Cref{tab:s_platforms} summarizes their configurations and execution parameters, and this section adds implementation details not covered in the main paper.

\subsection{Hardware and sensors}
\label{sec:s_sys_hw}
\textbf{Rokae.} All three cameras provide RGB and depth images at $640\times480$. The tactile sensors provide tactile images, deformation fields, and force and torque readings.

\textbf{Franka.} The two cameras also provide RGB and depth images at $640\times480$. The external wrench is referenced to the end-effector origin.

\subsection{Compliant execution (motion server)}
\label{sec:s_sys_ctrl}

\begin{figure}[t]
\centering
\includegraphics[width=0.92\colw]{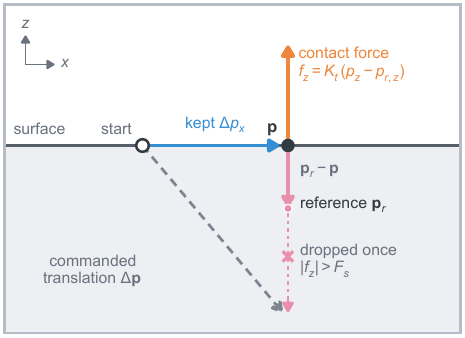}
\caption{Impedance tracking and force-aware motion clipping for one command that moves the end effector along a surface and into it, with the surface normal along the base $z$ axis. The $x$ component of the commanded translation $\Delta\mathbf{p}$ is kept. The $z$ component moves the reference $\mathbf{p}_r$ into the surface, which stretches the virtual spring and produces the contact force $f_z$, and it is dropped once $|f_z|$ exceeds the soft limit $F_s$.}
\label{fig:s_impedance}
\end{figure}

Sec.~3.4 of the main paper introduces the principle of compliant execution, and the implementation details are given here. Execution has two layers. An outer loop moves the reference pose of the end effector in small steps and clips each step by the measured force, and the impedance controller of the robot tracks this reference in the inner loop. The Rokae and Franka platforms execute one step every 50\,ms and 20\,ms, respectively. \Cref{fig:s_impedance} illustrates both layers for a command that moves along a surface and into it. The controller gains of the Franka platform are given in the released code.

\textbf{Impedance control.} The controller acts as a virtual spring and damper between the end effector and the reference,
\begin{equation}
\mathbf{F} = \mathbf{K}(\mathbf{x}_r-\mathbf{x}) - \mathbf{D}\dot{\mathbf{x}}, \qquad \boldsymbol{\tau} = \mathbf{J}(\mathbf{q})^{\top}\mathbf{F} + \boldsymbol{\tau}_c,
\end{equation}
where $\mathbf{x}$ and $\mathbf{x}_r$ are the measured and reference poses of the end effector, the rotational part of $\mathbf{x}_r-\mathbf{x}$ is the orientation error, $\mathbf{K}=\mathrm{diag}(K_t\mathbf{I}_3, K_r\mathbf{I}_3)$ and $\mathbf{D}$ are the stiffness and damping matrices, $\mathbf{J}$ is the Jacobian at the joint angles $\mathbf{q}$, and $\boldsymbol{\tau}_c$ compensates gravity and Coriolis forces. In free space, the end effector follows the reference as under position control. In contact, the surface holds the end effector back while the reference moves on, and the spring stretches. At rest, the external force on the end effector is then $\mathbf{f} = K_t(\mathbf{p}-\mathbf{p}_r)$, where $\mathbf{p}$ and $\mathbf{p}_r$ are the positions in $\mathbf{x}$ and $\mathbf{x}_r$. How far the reference lies beyond the surface therefore determines the contact force, and a position command becomes a force command.

\begin{algorithm}[t]
\caption{Force-aware motion clipping in the execution of one command}
\label{alg:s_clip}
\begin{algorithmic}[1]
\Require reference pose $\mathbf{x}_r$, commanded translation $\Delta\mathbf{p}$ and rotation $\Delta\mathbf{r}$, duration $T$, step period $\Delta t$, soft limit $F_s$, hard limit $F_h$
\State $N \gets \lceil T/\Delta t\rceil$, \ $\delta\mathbf{p} \gets \Delta\mathbf{p}/N$, \ $\delta\mathbf{r} \gets \Delta\mathbf{r}/N$
\For{$k = 1, \dots, N$}
  \State $\mathbf{f} \gets$ external force on the end effector in the base frame
  \If{$\|\mathbf{f}\| \geq F_h$}
    \State abort and cancel the rest of the batch
  \EndIf
  \State $\delta\tilde{\mathbf{p}} \gets \delta\mathbf{p}$
  \For{$i \in \{x, y, z\}$}
    \If{$|f_i| > F_s$ \textbf{and} $f_i\,\delta p_i < 0$}
      \State $\delta\tilde{p}_i \gets 0$
    \EndIf
  \EndFor
  \State $\mathbf{x}_r \gets \mathbf{x}_r \oplus (\delta\tilde{\mathbf{p}}, \delta\mathbf{r})$
  \State send $\mathbf{x}_r$ to the impedance controller and wait $\Delta t$
\EndFor
\State \Return $\mathbf{x}_r$ as the start of the next command
\end{algorithmic}
\end{algorithm}

\textbf{Force-aware motion clipping.} \Cref{alg:s_clip} gives the outer loop for one displacement or pose command. The command is divided into $N$ equal steps. Before each step, the loop reads the external force and aborts the batch if the force reaches the hard limit. It then clips the translation step axis by axis. The step along an axis is removed only when the force along that axis already exceeds the soft limit and the step would push further into the contact, that is, when $f_i\,\delta p_i<0$. The reference pose then moves by the clipped translation step and the unclipped rotation step, denoted by $\oplus$. A clipped part is discarded and not made up later, and the next command starts from the reference pose reached at the end of this one. Motion away from the contact and motion along the surface continue, so the agent can slide the tool over a surface while pressing on it, as in the surface search of insertion, in wiping, and in writing.

Together with impedance control, clipping stops the reference at the depth where the contact force reaches the soft limit (\cref{fig:s_impedance}), so the force along that axis stays close to $F_s$. The soft limit thus works as a force bound that the agent sets for each phase, low for search and high for seating, and the force is regulated within each step without waiting for the next decision.

The default soft limit and the motion range of a single command are listed in \cref{tab:s_platforms}. Notably, the agent can modify the soft limit through commands during task execution, and its prompt describes this command and when to raise or lower the limit.

\subsection{Observation interface (observation server)}
\label{sec:s_sys_obs}
The observation interface runs as a separate service. It reads all cameras, the tactile sensors, and the robot state continuously, keeps the latest sample of each source, and returns only the observations that the agent requests, together with an error for any sample older than 0.5\,s. The processing of each modality is described below for the Rokae platform. The Franka platform follows the same processing and has no tactile sensors.

\begin{table}[t]
\centering
\small
\setlength{\tabcolsep}{2pt}
\begin{minipage}{0.6\textwidth}
\caption{Configurations and execution parameters of the Rokae and Franka real-robot platforms.}
\label{tab:s_platforms}
\begin{tabular*}{\linewidth}{@{\extracolsep{\fill}}lll@{}}
\toprule
 & Rokae & Franka \\
\midrule
Arm & 7-DoF Rokae & 7-DoF Franka \\
Gripper & DH-Robotics AG-95 & Franka gripper \\
Wrist camera & Intel RealSense D405 & Intel RealSense D405 \\
Head camera & Orbbec Gemini 335L & None \\
Third-view camera & Other wrist D405 & Intel RealSense D435i \\
Image resolution & $640\times480$ & $640\times480$ \\
Tactile sensor & Daimon DM-Tac W2 & None \\
\midrule
Step rate & 20\,Hz & 50\,Hz \\
Max translation & 0.15\,m & 0.15\,m \\
Max rotation & $90^\circ$ & $90^\circ$ \\
Max linear speed & 0.035\,m/s & 0.025\,m/s \\
Max angular speed & 0.08\,rad/s & 0.08\,rad/s \\
Command duration & 0.05--120\,s & 0.02--120\,s \\
Default soft limit & 10\,N & 10\,N \\
Hard limit & 50\,N & 50\,N \\
\bottomrule
\end{tabular*}
\end{minipage}
\end{table}

\textbf{Vision.} Two Intel RealSense D405 cameras on the wrists and an Orbbec Gemini 335L camera on the head stream RGB and depth images at $640\times480$ and 30\,Hz. For each camera, the interface aligns the depth image to the color image, converts depth to meters, and returns the color image as a PNG file and the depth map as an array.

\textbf{Proprioception.} The interface reads the real-time state of the robot at 20\,Hz. It returns the end-effector pose $[x, y, z, r_x, r_y, r_z]$ in the robot base frame, in meters and XYZ Euler angles in radians, the seven joint angles in radians, the gripper opening, and the status of the motion server, which tells whether a command is still running. These values are passed on without transformation.

\textbf{Force.} The robot controller reports the external wrench at the flange, ${}^{F}\mathbf{f}$ and ${}^{F}\mathbf{m}$, in the flange axes. The interface calibrates a baseline by averaging over 1\,s while the robot is stationary and free of contact, at startup and whenever the agent requests it, and subtracts it to obtain ${}^{F}\tilde{\mathbf{f}}$ and ${}^{F}\tilde{\mathbf{m}}$. It then moves the reference point of the moment from the flange to the tool center point (TCP) and expresses the wrench in world axes,
\begin{align}
\mathbf{f}_{\mathrm{obs}} &= \mathbf{R}_{WB}\,\mathbf{R}_{BF}\,{}^{F}\tilde{\mathbf{f}}, \\
\mathbf{m}_{\mathrm{obs}} &= \mathbf{R}_{WB}\,\mathbf{R}_{BF}\,\big({}^{F}\tilde{\mathbf{m}} - {}^{F}\mathbf{p}_{T}\times{}^{F}\tilde{\mathbf{f}}\big),
\end{align}
where ${}^{F}\mathbf{p}_{T}$ is the position of the TCP in the flange frame from the tool calibration, $\mathbf{R}_{BF}$ is the orientation of the flange in the base frame, and $\mathbf{R}_{WB}$ is a fixed rotation from the base axes to world axes aligned with the workspace. Taking the moment about the TCP means that a force acting at the TCP produces no moment, so a nonzero moment indicates a load away from the TCP, such as a connector pressed off its axis. On the Franka platform, the wrench is the external wrench estimated by the robot and goes through the same steps.

\textbf{Tactile sensing.} Each finger of the gripper carries a Daimon DM-Tac W2 visuotactile pad. For each pad $j$, its SDK provides a tactile image, a deformation field with the in-plane displacement and the depth at each point, and a wrench $(\mathbf{f}_j, \mathbf{m}_j)$ in the axes of the pad. The interface subtracts from the wrench and the deformation a baseline that is calibrated on the agent's request after each new grasp. It then rotates the two pad wrenches into common axes and sums the forces, and the agent sums the torques as instructed in the system prompt,
\begin{align}
\mathbf{f}_{\mathrm{tac}} &= \mathbf{R}_{WB}\,\mathbf{R}_{BF}\sum_{j\in\{L,R\}}\mathbf{R}_{FS_j}\big(\mathbf{f}_j-\bar{\mathbf{f}}_j\big), \\
\mathbf{m}_{\mathrm{tac}} &= \mathbf{R}_{WB}\,\mathbf{R}_{BF}\sum_{j\in\{L,R\}}\mathbf{R}_{FS_j}\big(\mathbf{m}_j-\bar{\mathbf{m}}_j\big),
\end{align}
where $\mathbf{R}_{FS_j}$ is the fixed rotation from the axes of pad $j$ to the flange axes, and $\mathbf{R}_{WB}\mathbf{R}_{BF}$ is the same rotation to world axes as for the force. The pads measure the load where the connector is held, so a bending of the connector against the socket appears in $\mathbf{m}_{\mathrm{tac}}$. Because the SDK does not specify the unit of the torque, the agent uses $\mathbf{m}_{\mathrm{tac}}$ as a qualitative signal of the bending load and judges a correction by whether this load decreases.

\subsection{Command interface}
\label{sec:s_sys_act}
Besides the displacement command introduced in the main paper, the command interface provides a pose command and a gripper command. A displacement command specifies a relative translation vector and a rotation vector in axis-angle form. A pose command gives a target position and orientation, and a gripper command gives an opening between 0 (closed) and 1 (open). A command batch can contain at most 16 commands, which are executed in order. Batches rarely reached this limit in our experiments.

\section{Textual Skills}
\label{sec:s_skill}
In each trial, the agent reads one system prompt and one task prompt. The system prompt is shared across all tasks, and each task category has its own task prompt. All 15 connector types share one insertion prompt. The wording template of the opening prompt is ``Without using any skill in the system, please read the instruction in the system prompt and task prompt under [directory], then start the task. [Minor hints, for cables with special requirements.]'' The final placeholder is filled with a one-sentence hint specific to the cable or socket. For example, the hint may describe a curved contact surface on the socket housing, a small grasping area that makes the cable slip easily, or an unusual cable shape that blocks one camera view and requires the agent to rely more on the other.

\begin{figure*}[t]
\centering
\includegraphics[width=\textwidth]{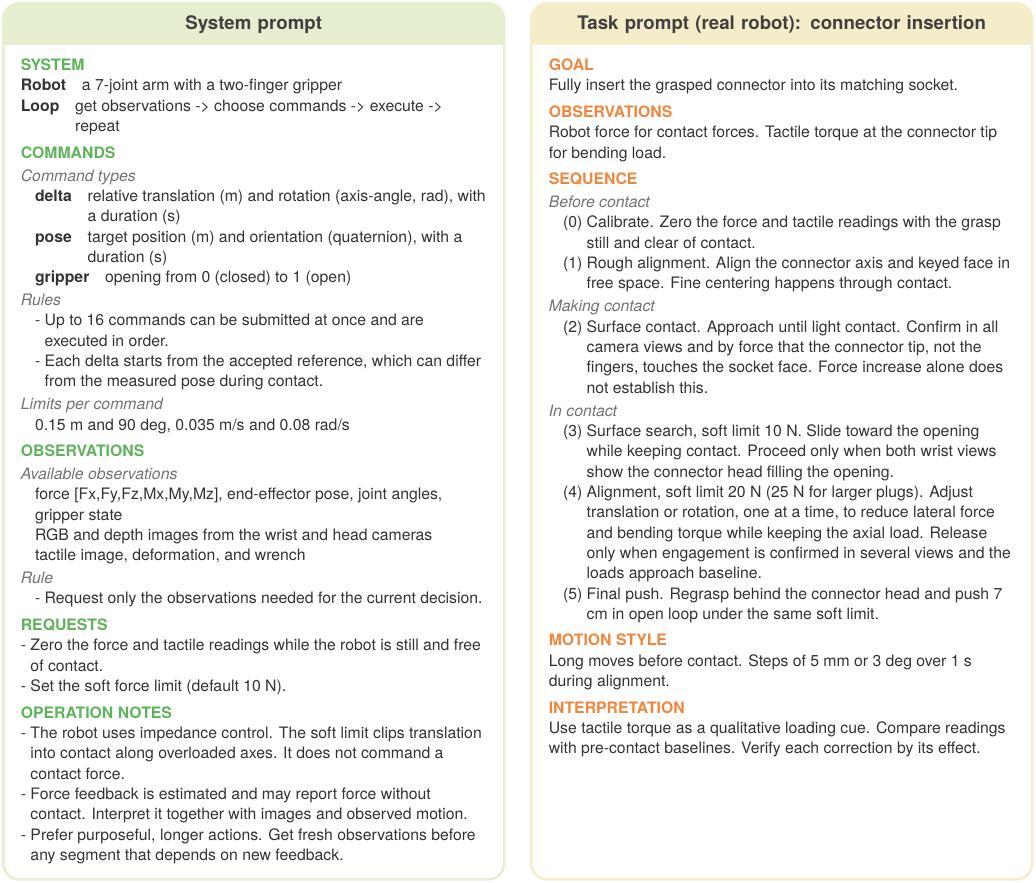}
\caption{Summary of the system prompt (left) and the connector insertion task prompt (right), condensed from the prompts used in the experiments.}
\label{fig:s_prompts}
\end{figure*}

\Cref{fig:s_prompts} summarizes the system prompt and the insertion task prompt. The task prompts for other tasks are provided at the end of this appendix, before references. Because the complete prompts are lengthy, we provide them in the code released with the paper.

The system prompt explains to the agent how to use Contact Harness. Its content is independent of the task and corresponds to the left panel of \cref{fig:s_prompts}. It covers the following items.
\begin{itemize}
    \item It describes the robot and the decision loop, in which the agent obtains observations, selects commands, and waits for execution to finish in each round.
    \item It defines the motion commands, including the parameters of the displacement, pose, and gripper commands, the magnitude and speed limits of a single command, and the rules for batch submission.
    \item It provides the command for acquiring observations and lists all observations that can be requested. The agent is asked to request only the observations needed for the current decision.
    \item It describes additional commands for zeroing the force and tactile readings and adjusting the soft limit.
    \item The prompt ends with operating rules shared by all tasks.
\end{itemize}

\vspace{5pt}
A task prompt explains how to complete one category of tasks. The right panel of \cref{fig:s_prompts} uses connector insertion as an example. The prompt first gives the task goal, the initial state, and the completion condition, and then specifies how to use the two readings. The main body consists of six steps executed in order.
\begin{itemize}
    \item The first two steps take place before contact. The agent zeroes the readings and then uses vision to roughly align the connector with the socket in free space.
    \item In the third step, the agent establishes surface contact and uses all available evidence to confirm that contact is at the intended surface.
    \item The next two steps are carried out under contact. The agent first slides along the surface toward the opening under a sustained contact force. Once the plug is aligned with the opening, the agent adjusts translation or rotation based on the lateral force and bending torque to reduce their magnitudes while advancing the connector.
    \item In the last step, the agent regrasps the connector and completes the push if the plug is not fully inserted visually.
\end{itemize}
Each step states the condition for moving to the next step. These conditions combine images, force readings, and tactile readings.

After describing the main procedures, the task prompts also highlight important considerations to reduce ambiguity in action selection. Except for the opening prompt, these prompts contain no rules written for specific connector types. They can include guidance on how to handle connectors of different sizes, how to proceed when a connector has no cable, or when to use a different soft limit during alignment. However, the agent itself evaluates the observations and makes all behavioral decisions.

With the above, all 15 connector types therefore share one insertion task prompt. The task prompts for whiteboard wiping, writing, peeling, and pivoting have the same structure and differ in their steps and contact targets.

\section{Real-Robot Experiments}
\label{sec:s_real}
This section supplements the real-robot experiments in Sec.~4.2 of the main paper. The first part analyzes failures in connector insertion. The second part uses USB-A as an example and counts how often the agent requests each observation modality during each phase of insertion. The third part visualizes the execution of each task.

\subsection{Failure analysis}
\label{sec:s_fail}
Of the 150 insertion trials reported in Table~1 of the main paper, 14 failed across 8 connector types. Mini DisplayPort and Micro USB accounted for 3 failures each, USB-B and DisplayPort for 2 each, and Ethernet, Lightning, 4-Pin Connector, and VGA for 1 each. The main causes are described below.

\textbf{Plug slips between the fingers.} For small connectors with small housings, the contact area between the housing and the finger pads is too small to provide enough friction to prevent slipping. Pushing the connector against the socket surface can then cause it to rotate between the fingertips. The gripping force is also limited because grasping the connector too tightly would damage the membrane of the visual-tactile sensor.

\textbf{Insufficient camera resolution.}\label{sec:s_res_microusb} Very small connectors occupy only a few pixels in the camera image. For Micro USB, the plug is much smaller than the fingertip, and the image is too blurry for the agent to judge whether the plug rests on the correct contact surface or has jammed.

\textbf{Occluded views.} When the plug is small, the gripper fingers can block it from view in the other cameras, leaving the agent without an alternative view of the contact.

\textbf{Model service stalls.} The model service was unstable during the experiments, with disconnections and long periods without a response. After some disconnections, the session reconnected automatically and the trial continued. In other cases, the disconnection caused the trial to fail. We did not count these two outcomes separately. These service interruptions lengthened task execution and were the main source of variation in execution time in Table~1 of the main paper.

\begin{figure}[t]
\centering
\begin{minipage}{\colw}
\includegraphics[width=\linewidth]{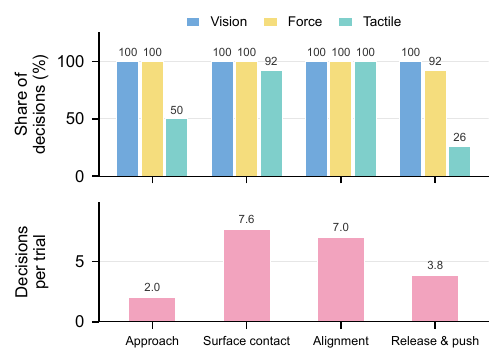}
\caption{Observations obtained by the VLM in each phase over 10 real-robot USB-A insertion trials. The upper part shows the share of decisions preceded by each type of signal, and the lower part shows the average number of decisions per trial.}
\label{fig:evidence}
\end{minipage}
\end{figure}

\subsection{Observation requests in each phase}
\label{sec:s_real_evidence}
USB-A belongs to the easiest group in Table~1 of the main paper. It is large enough to see clearly in the camera image, has simple alignment requirements, fits relatively loosely, and rarely jams. We analyze the logs of 10 real-robot USB-A insertion trials to examine which observation modalities the VLM uses for decisions in each phase of insertion. The results are shown in \cref{fig:evidence}.

Following the task prompt described above, we divide each trial into four phases, namely approach, surface contact, alignment, and release and push. The boundaries between successive phases are marked by the first detected contact force, the increase in the soft limit, and the first opening of the gripper, respectively. The upper panel shows the proportion of decisions in each phase for which the VLM requests each modality. Camera images are requested for every decision across all four phases. Wrist force readings are requested for every decision during approach, surface contact, and alignment, and for 92\% of release and push decisions. Tactile readings are requested for 50\% of approach decisions, 92\% of surface contact decisions, 100\% of alignment decisions, and 26\% of release and push decisions. The lower panel shows the average number of decisions per trial, with 2.0 for approach, 7.6 for surface contact, 7.0 for alignment, and 3.8 for release and push.

Vision supports decisions throughout insertion, from roughly aligning the connector during approach to assessing insertion depth during the final push. Wrist force readings help the VLM assess whether contact has occurred and how the load changes as the connector advances. Tactile readings are used mainly during surface contact and alignment, when the VLM assesses connector bending to decide whether to adjust its position or orientation. Their use drops after the gripper opens because the fingertips no longer hold the connector. Surface contact and alignment also account for 14.6 of the 20.4 decisions per trial on average, or more than 70\%.

\subsection{Execution snapshot}
\label{sec:s_real_show}
\begin{figure*}[t]
\centering
\includegraphics[width=\textwidth]{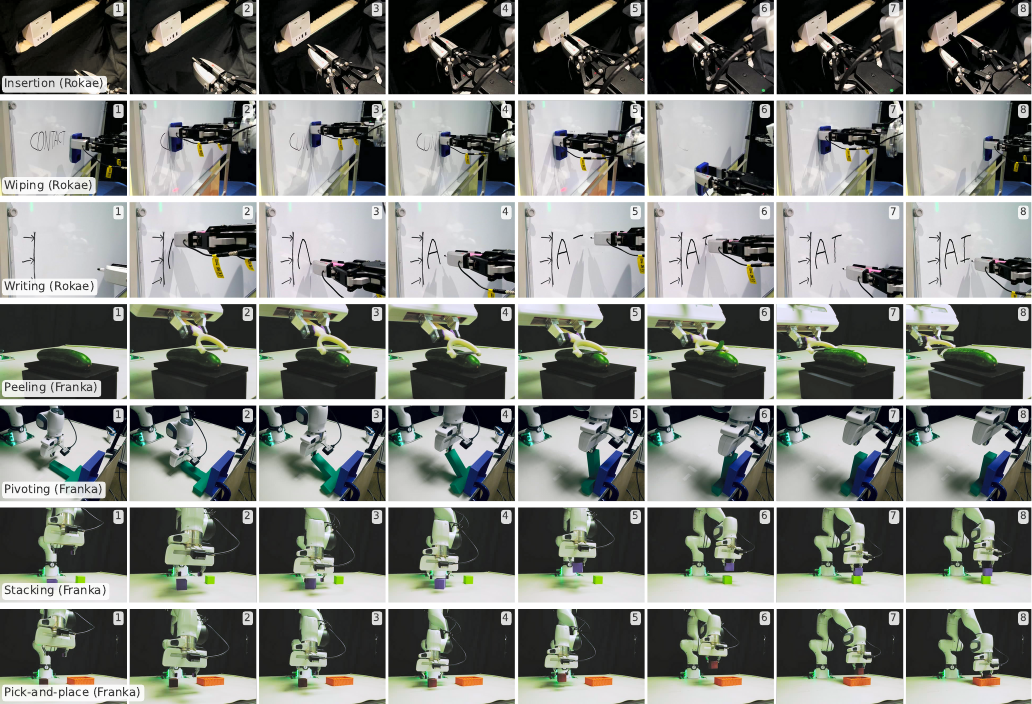}
\caption{Execution of the seven real-robot tasks. Each row shows one trial of one task with eight keyframes in temporal order from left to right. The first three rows are on the Rokae platform and the last four on the Franka platform.}
\label{fig:s_real_keyframes}
\end{figure*}

\Cref{fig:s_real_keyframes} shows the execution of one trial for each of the seven real-robot tasks. Each row corresponds to one trial of one task. The frames are taken from the third-view recording and show eight keyframes in temporal order from left to right. The first three rows are connector insertion, whiteboard wiping, and writing on the Rokae platform, and the last four rows are peeling, pivoting, stacking, and pick-and-place on the Franka platform.

\section{Simulated Benchmark}
\label{sec:s_sim}
This section supplements the simulated benchmark in Sec.~4.3 of the main paper. \Cref{sec:s_sim_tasks} describes the construction of the six simulated tasks, and \cref{sec:s_eval_sim} defines the success criteria and explains how each metric is computed. \Cref{sec:s_res_base} describes how we reproduce each VLM-based baseline and what support we provide for it. \Cref{sec:s_bench_vla} presents the benchmark with its demonstrations and the VLA baselines trained on them. \Cref{sec:s_res_ours} presents the results of Contact Harness and compares the agent models, and \cref{sec:s_fail_base} analyzes baseline failures by method and by task. \Cref{sec:s_res_abl} gives additional ablation results.

\subsection{Simulated task construction}
\label{sec:s_sim_tasks}
This subsection describes the construction of the six simulated tasks. It first covers the environment shared by all tasks and then the scene, initial state, and randomization of each task.

\subsubsection{Environment Specifications}

\textbf{Simulator.} All tasks run in MuJoCo 2.3.2 and robosuite 1.4.0. The table height is 0.9\,m in Wipe and 0.907\,m in the other tasks. The physics time step is 2\,ms in Wipe, Place, and Stack, 1\,ms in Pivot, and 0.5\,ms in Insert and Write. Except for Wipe, all tasks use elliptic friction cones with \texttt{impratio} set to 100 and 50 noslip iterations to prevent grasped objects from rotating or slipping in the hand. The physics simulation is synchronized with real time.

\textbf{Robot and control.} The robot is the robosuite Panda arm on its default mount. Wipe uses a wiping pad, and the other tasks use the Panda parallel gripper. A Cartesian impedance controller drives the arm with a translational stiffness of 400\,N/m, a rotational stiffness of 50\,N$\cdot$m/rad, and a damping ratio of 0.9. The controller includes gravity and Coriolis compensation. The gripper fingers use position control with a gain of $2 \times 10^4$\,N/m, a grip force limit of 70\,N per finger, and a maximum finger speed of 0.05\,m/s.

\textbf{Contact Harness motion interface.} Although the simulation uses a Franka Panda arm, the motion and observation services on the agent side use the same programs as on the real-world Rokae platform. The only change is the replacement of the Rokae kinematic model with that of the Panda arm. Commands therefore follow the execution parameters in the Rokae column of \cref{tab:s_platforms}. The workspace is a cube extending $\pm 0.30$\,m along each axis from the end-effector position at the start of the first motion.

\begin{figure*}[t]
\centering
\includegraphics[width=\textwidth]{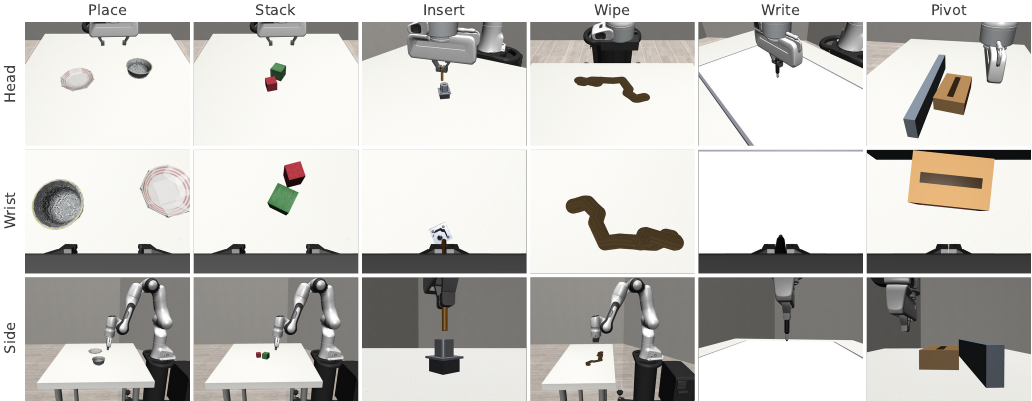}
\caption{The scenes of the six simulated tasks at the start of a run. Each column is a task, and the rows from top to bottom are the head camera, the wrist camera, and the side camera, which are the three cameras the agent can request.}
\label{fig:s_sim_intro}
\end{figure*}
\textbf{Observations.} The observation interface keeps the three camera names used on the real Rokae robot. The head camera corresponds to a fixed camera in front of the scene, the left wrist camera to the camera mounted on the hand, and the right wrist camera to a fixed camera at the side of the scene. All three cameras provide $640 \times 480$ RGB images and depth maps in meters. The robot state includes the joint angles, end-effector pose, gripper opening, and contact force. The contact force computation is described in \cref{sec:s_eval_sim}. At startup, the observation service uses the average contact force over 1\,s as the zero point, and the agent can zero the readings again at any time. \Cref{fig:s_sim_intro} shows the images from all three cameras at the start of each task.

\textbf{Tactile sensing.} Tactile observations are synthesized from contacts on the finger pads. The simulator provides the position and force of each contact point on each finger pad, and the observation service converts these values into observations in the same format as those from the real sensor. The tactile area measures $24 \times 18$\,mm. The deformation field is a superposition of Gaussian indentations centered at the contact points, each with a standard deviation of 2.5\,mm. Normal force produces an indentation depth of 0.08\,mm per newton, and tangential force produces a displacement of 0.04\,mm per newton. The tactile image is rendered from the deformation field. The tactile-based force reading is the sum of the contact forces plus noise with a standard deviation of 0.02\,N. The wipe task has no tactile observations because its wiping pad has no finger pads.

\subsubsection{Task Specifications}

\textbf{Insert.} Insert reproduces the PegInsert task of Isaac Lab Factory in MuJoCo with the original meshes, masses, friction coefficients, randomization ranges, and success criterion. The peg has a diameter of 7.986\,mm and a length of 50\,mm. The block with the hole has a $40 \times 40$\,mm footprint, the hole is 25\,mm deep, and its opening has a 1\,mm chamfer. The hole diameter takes the nominal value of 8.1\,mm from the Isaac Lab Factory configuration instead of the 9.0\,mm bore of the original mesh, which reduces the radial clearance from about 0.5\,mm to about 0.06\,mm.

The hole position is centered 0.6\,m in front of the robot base at a height of 0.05\,m and is randomly offset by $\pm 5$\,cm along each of the three directions. The hole orientation about the vertical axis is sampled uniformly over the full circle. At the start of each run, the peg is already held in the gripper above the hole, with a horizontal offset of $\pm 2$\,cm and a hand yaw about the vertical axis within $\pm 45^\circ$. The lower end of the peg is about 0.2 to 2.8\,cm above the top face of the block, and the position of the peg in the hand has a random offset of $\pm 3$\,mm.

The task is harder than the Factory original in three ways. First, the clearance is tighter, as described above. Second, the grasped peg is subject to gravity, which Factory disables. Third, press-fit resistance is added to reflect the insertion resistance of real connectors. Once the peg is inserted by 3\,mm, it must overcome a friction force of 15\,N to advance further, all the way to the hole bottom. The value of 15\,N lies between the 10\,N soft limit for search and the 20\,N soft limit for insertion in the real-robot task prompt.

\textbf{Place.} Place requires placing a bowl on a plate. The scene is based on the robosuite Stack environment, with the bowl and plate meshes from LIBERO scaled to about 11\,cm and 14\,cm wide. The bowl is wider than the 8\,cm maximum opening of the gripper and can only be grasped by its rim. The positions of the plate and the bowl are sampled uniformly within a $10 \times 8$\,cm region and a $10 \times 6$\,cm region on opposite sides of the table center, 18 to 32\,cm apart laterally, and their orientations about the vertical axis are sampled uniformly over the full circle.

\textbf{Stack.} Stack is the robosuite Stack task, in which a red cube with 4\,cm sides is placed on a green cube with 5\,cm sides. The cube positions are sampled uniformly within $\pm 8$\,cm of the table center without overlap, and their orientations about the vertical axis are sampled uniformly over the full circle. The scene and the success criterion follow robosuite, and the only changes are the table height, the gripper servo, and the contact solver parameters.

\textbf{Wipe.} Wipe uses the Wipe environment of robosuite. The wiping pad is the robosuite WipingGripper, rigidly mounted on the end flange, with no fingers and a rectangular wiping surface of $12 \times 5$\,cm. The table carries 100 dirt markers, each being a disc with a radius of 2\,cm. The markers are laid out 5\,mm apart along a continuous random path that stays within the central 60\% of the table. The initial joint angles are the default robosuite Panda pose plus Gaussian noise. This default pose tilts the hand by about $7^\circ$. At the start of each run, including Contact Harness and the baselines, we rotate the hand so that the wiping surface is parallel to the table, keeping its position unchanged.

\textbf{Write.} Write is adapted from the DrawSVG task of ManiSkill3. The target word is ROBOT with a letter height of 5\,cm, and the task does not specify where to write. The whiteboard is $0.70 \times 0.45$\,m and lies flat on the table with the writing surface facing up. The board center is 0.55\,m in front of the robot base and varies with the seed within $\pm 1$\,cm front to back and $\pm 3$\,cm left to right.

The pen tip is a sphere with a radius of 2.5\,mm, and its contact with the board is set to be soft to model a felt tip. At the start of each run, the pen is already held in the gripper, perpendicular to the board, with the tip extending 6\,cm beyond the fingertips, $4 \pm 1$\,cm above the board, and within $\pm 2$\,cm of the board center horizontally.

Ink is generated from force. When the normal force between the tip and the board reaches 0.5\,N, an ink dot is left at the contact point and displayed as a disc with a diameter of 3.6\,mm. The tip is marked as damaged when the instantaneous normal force on it exceeds 20\,N. The mark is only a record and is not a success criterion.

\textbf{Pivot.} Pivot is a new task proposed in this paper, in which the robot pushes a box lying flat on the table against a low wall until the box stands on its narrow side. The wall is $3 \times 40 \times 7$\,cm and fixed on the table. The wall face is $0.64 \pm 0.02$\,m in front of the robot base, and the lateral position of the wall varies within $\pm 3$\,cm.

The box has nominal dimensions of $10 \times 16 \times 5$\,cm in depth, width, and thickness, and each side varies with the seed within $\pm 5$\,mm. Its depth and width both exceed the 8\,cm maximum opening of the Panda gripper, so it cannot be grasped directly. The box mass is sampled uniformly between 0.05\,kg and 0.30\,kg, its yaw about the vertical axis is within $\pm 8^\circ$, its nearest corner is 0 to 2\,cm from the wall, and its lateral position is within $\pm 4$\,cm. The friction coefficient between the box and the table is sampled between 0.3 and 0.6, that between the box and the wall between 0.2 and 0.45, and that between the box and the fingers is 1.0.

At the start of each run, the fingers are closed and point downward, with the fingertips 6\,cm in front of the near face of the box, $10 \pm 1$\,cm above the table, and within $\pm 2$\,cm of the box center laterally.

The tilt angle of the box is the angle between its largest face and the table. It is 0 when the box lies flat and $90^\circ$ when the box stands on its narrow side. The pushing force of the robot on the box is the sum of the magnitudes of all contact forces between the box and the parts of the robot.

\begin{figure*}[t]
\centering
\includegraphics[width=\textwidth]{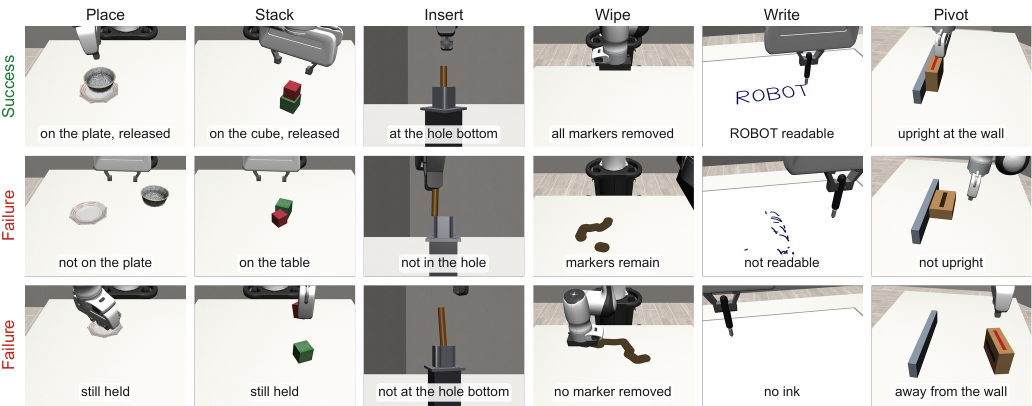}
\caption{Final scenes of successful and failed runs on the six simulated tasks. The top row shows a run that meets the success condition of each task. The two rows below show failed runs that violate different parts of the condition. In Place and Stack, the object is left beside its target or is still held by the gripper. In Insert, the peg rests beside the hole or stops in the opening above the hole bottom. In Wipe, some or all of the dirt markers remain. In Write, the strokes cannot be read or the pen leaves no ink. In Pivot, the box stays flat or stands upright away from the wall.}
\label{fig:s_success}
\end{figure*}
\textbf{Task prompts and random seeds.} The task prompts of Insert, Wipe, Write, and Pivot come from the corresponding real-robot task prompts. Because the simulated environments are simpler, these prompts are simplified and the constraints specific to the real robot are removed. Summaries of the prompts are in \cref{sec:s_prompts}. Each task uses 10 random seeds shared by all methods. The seed determines all random quantities listed above, including the initial poses of the objects and the end effector, the layout of the dirt markers in Wipe, and the box dimensions, mass, and friction coefficients of the contact surfaces in Pivot. With the same seed, all methods start from exactly the same initial state.

\subsection{Success criteria and metrics}
\label{sec:s_eval_sim}
This subsection defines the success criteria and the four metrics, success rate, time, number of decisions, and contact force, reported in all simulation tables, which are Tables~4 and~5 of the main paper and \cref{tab:s_agents,tab:s_prompt_effect,tab:s_ablation}.

\textbf{Rollout termination conditions.} The simulator checks the success condition of the task every 20\,ms. For the VLA policies, a rollout of Insert, Wipe, Pivot, Place, or Stack counts as successful the first time the condition holds, and the simulation terminates at that moment. A Write rollout runs until its step budget is used up. For Contact Harness and the VLM-based baselines, the agent decides when to stop the rollout, but the success label is given by the success conditions and not by the agent.

\textbf{Success conditions.} \Cref{fig:s_success} shows a successful final scene and two failed ones for each task.
\begin{itemize}
    \item Place requires that the bowl touches the plate, that the bowl center is above the plate and within 6\,cm of the plate center horizontally, and that the bowl is not held by the gripper.
    \item Stack follows the robosuite Stack criterion, in which the red cube is off the table, touches the green cube, and is not held by the gripper.
    \item Insert follows the Isaac Lab Factory PegInsert criterion, in which the horizontal offset between peg and hole is below 2.5\,mm and the peg is within 1\,mm of the hole bottom.
    \item Wipe requires all 100 dirt markers to be removed. The robosuite removal rule is lenient and does not require the pad to press on the table. However, across all runs, only 3 markers were removed at a contact force below 0.05\,N.
    \item Pivot requires four conditions at once. The tilt angle of the box is at least $80^\circ$, the lowest corner of the box is less than 4\,mm above the table, its nearest corner is at most 3\,cm from the wall, and the box stays still for 2\,s without robot contact.
    \item Write is judged by a person for Contact Harness and the VLM-based baselines. A run succeeds if the word ROBOT can be read, with no requirement on font, size, or position. The Write runs of the VLA policies are judged by GPT-6 Astra with the same criterion, from three views of the whiteboard at the end of the rollout.
\end{itemize}

\textbf{Contact force.} The contact force is the resultant force between the tool and everything it touches. The tool is the wiping pad in Wipe and the gripper with its fingers in the other tasks. The grasped peg, the pen, and the pushed box are not part of the tool, and their contact with the environment counts through the force they pass to the gripper. We report the sustained peak contact force, which is the largest force held for at least 0.2\,s during a run.

\textbf{Time and number of decisions.} Time is wall-clock time, from when the simulator and the services are ready until the run ends. For Contact Harness and the VLM-based baselines, success must be reached within 20 minutes for Place, Stack, and Insert and within 30 minutes for Wipe, Write, and Pivot. For the VLA policies, a rollout is capped by the number of action steps, with 350 for Place, 400 for Pivot, 600 for Stack, Insert, and Wipe, and 1800 for Write, which corresponds to 35\,s to 180\,s.

The number of decisions is the number of motion command calls by the agent for Contact Harness, the number of generated code blocks for CaP-X, fixed at 1 for GenCHiP, the number of agent calls for Show-Harness, and the number of policy inferences for the VLA policies.

\textbf{Statistics in the tables.} The success rate is computed within each task and then averaged over tasks. Time and the number of decisions are averaged over the successful runs of each task and then over the tasks with at least one successful run. In Table~4 of the main paper, these two averages leave out Write, as explained in the next paragraph. Force is the median of the sustained peak contact force over all runs of each task, successful or not, averaged over tasks.

\textbf{Why Write is treated separately.} Write is the only task whose success cannot be decided by a simple programmatic check, since reading the word ROBOT takes the judgement of a person or an agent. Running such a judgement after every step of a VLA policy is impractical. A VLA policy therefore does not stop by itself in Write, and each evaluation runs for the full 1800 steps before GPT-6 Astra judges the writing. In contrast, the agent methods end by themselves once the word is written, as the agent can judge the completeness of the writing at any time. Time and the number of decisions are thus not comparable between the two kinds of methods on this task, so Table~4 of the main paper excludes Write from these two metrics.

\textbf{Force-related criteria.} Comparisons with baselines in Table~4 use no force-related success criteria, and the peak contact force is reported for analysis. The ablation, however, adds the force criteria below on top of the success conditions. In all tasks, the sustained peak contact force should not exceed 50\,N. In Wipe, all 100 markers should be removed as before, and at least 80 of them should be removed while the contact force is at least 5\,N. The margin of 20 markers allows for the sampling interval of the recording and for markers removed at the edges of the pad. In Write, the normal force on the pen tip should not exceed 20\,N. In Pivot, the sustained pushing force of the robot on the box should not exceed 14\,N, a bound that follows the box flipping task of ImplicitRDP~\cite{chen2026implicitrdp}.

\subsection{VLM-based baselines}
\label{sec:s_res_base}
The VLM-based baselines use the same simulator, the same initial states (random seeds), and the same model (GPT-6 Astra with high reasoning effort) as Contact Harness, with the time limits and success criteria of \cref{sec:s_eval_sim}. Success depends only on whether the task itself is completed, without force criteria. Each baseline is described below together with the support we provide to it, none of which exists in the original method. \Cref{sec:s_fail_base} analyzes the failure cases.

\subsubsection{Show-Harness}
\label{sec:s_res_sh}
\textbf{Reproduction.} We use the public real-robot code of Show-Harness for the Franka and run it zero-shot. At every step, Show-Harness selects one discrete action that moves the end effector along one coordinate axis, and we use its smallest step of 1\,cm. The real robot is replaced by the simulated one, with end-effector speed limits of 0.25\,m/s and 1\,rad/s. Show-Harness keeps its own two cameras (front and wrist) and its own impedance parameters. It does not use the compliant execution or the soft limit of Contact Harness, and its observations contain no force or tactile information.

The robot is in contact with the environment most of the time, and contact in simulation is stiff. To keep frequent protective stops under large force from disrupting normal execution, a motion first backs off when the contact force stays above 50\,N for 0.2\,s, that is, its target pose is pulled back so that it lies less deep inside the object. The motion is aborted only if the force remains above 50\,N for another 0.2\,s.

\textbf{Support.} The released prompt of Show-Harness forbids back-and-forth motion, discourages lateral adjustment, and forbids pressing down once the bottom is reached. This is reasonable, as Show-Harness is designed for free-space pick-and-place, where such motions are inefficient. In contact-rich tasks, however, back-and-forth lateral adjustment and pressing under a constant force are essential for success. We therefore rewrite the control prompt to allow these motions and add manipulation strategies to the task descriptions of the four contact tasks.

We also give Show-Harness the contact height of every scene. Its safety layer keeps the end effector above a height that the original method calibrates on the real robot. We set this height to where the tool just reaches the surface, namely the fingertips on the top face of the block in Insert, the pen tip pressing the board with 1\,N in Write, and the wiping surface on the table in Wipe. In Write and Wipe, a downward move therefore ends in the light contact that these two tasks need, and Show-Harness does not have to find the surface from images. In Insert, the bound lies at the top face of the block and does not limit the push on the peg.

Because our success conditions can be stricter than those of the original setting, we further add a completion check. If Show-Harness ends a run before the success condition is met, it is informed and continues, except in Write. Neither the original method nor Contact Harness has such a check. The step limit is 200 and is reset after each completion check, compared with 50 steps per run in the original evaluation.

\subsubsection{CaP-X}
\label{sec:s_res_capx}
\textbf{Reproduction.} We use the public real-robot code of CaP-X for the Franka. The real robot is replaced by the simulated one, with end-effector speed limits of 0.25\,m/s and 1\,rad/s. In each round, CaP-X generates a code block and executes it open loop. After execution, a visual differencing model compares the images before and after, and the next round begins. The model that generates code does not see images directly, which is the default CaP-X setting. Each episode allows one attempt. The real-robot interface has a single third-view camera, and this is CaP-X in the main paper. CaP-X + w.cam adds the wrist camera of Contact Harness to examine the effect of the number of cameras, following the way CaP-X handles the wrist camera in LIBERO. The visual differencing model and object localization then use both views. CaP-X + our prompt skill adds the content of the Contact Harness task prompt to the task description, whereas the original method only has the task goal.

\textbf{Support.} Given that the tool length shipped with CaP-X is longer than the tool in simulation, we calibrate the tool tip at the start of each run. The original method has no force protection. We add the same protection as for Show-Harness (\cref{sec:s_res_sh}), and execution can continue after an aborted motion. We further add a completion check. If CaP-X ends a run before the success condition is met, it is informed and continues, except in Write. Neither the original method nor Contact Harness has such a check. The time limit is 20 or 30 minutes, compared with 1000\,s per run in the original setting. The budget of code blocks stays at the CaP-X default of 12 per attempt.

\subsubsection{GenCHiP}
\label{sec:s_res_genchip}
\textbf{Reproduction.} GenCHiP has no public code, and we reimplement it following Listing~1 and Secs.~IV and V of its paper. The agent is called once and returns a program composed of compliant moves, which is executed open loop with no further agent calls. The agent does not see images (see the next paragraph). The control interface follows the paper and includes Cartesian admittance moves with termination conditions on force and position error, opening and closing the gripper, reading the end-effector pose, and object pose estimation. Force protection is the same as for Show-Harness (\cref{sec:s_res_sh}). The prompt consists of the four parts in the paper, namely the task description, the control interface description, hints, and rules, without task examples. GenCHiP + our prompt skill adds the content of the task prompt.

\textbf{Support.} The original GenCHiP obtains object poses from a specially trained detector with a reported error of at most 4\,mm. This detector is not available, so we give the agent the ground-truth object position from the simulator plus an error of 0 to 4\,mm in a random direction, which mimics the reported detection error. The interface also directly provides the positions of the hole opening, the lower end of the peg, and the pen tip, as well as the extent of the dirt markers not yet removed, none of which the original method has. Because the agent writes the program without seeing the scene, the interface description also states the opening direction of the fingers and the offset of the fingertips from the tool point, which it would otherwise have to guess.

\subsection{Benchmark and VLA baselines}
\label{sec:s_bench_vla}
The six simulated tasks form a new benchmark for contact-rich manipulation. It serves both training-free agents, which use the observation and motion interfaces, and learned policies, which are trained on its demonstrations. \Cref{sec:s_bench} describes the benchmark and its demonstrations, and \cref{sec:s_res_vla} describes the VLA baselines trained on them.

\subsubsection{Contact-rich manipulation benchmark}
\label{sec:s_bench}
Insert, Wipe, Write, and Pivot require sustained and regulated contact, and Place and Stack serve as contact-free references. We name the benchmark Contact-4 after its four contact-rich tasks. All tasks run in one MuJoCo simulator with the same robot, controller, and sensors (\cref{sec:s_sim_tasks}). Notably, we deliberately include the two contact-free tasks to test whether a model or an algorithm that works on contact-rich tasks can also handle free-space motions without modification. This matters because contact-rich tasks require fine and mostly slow adjustments, whereas contact-free tasks use large and usually fast motions, so a method may have to trade one off against the other.

\begin{figure*}[t]
\centering
\includegraphics[width=\textwidth]{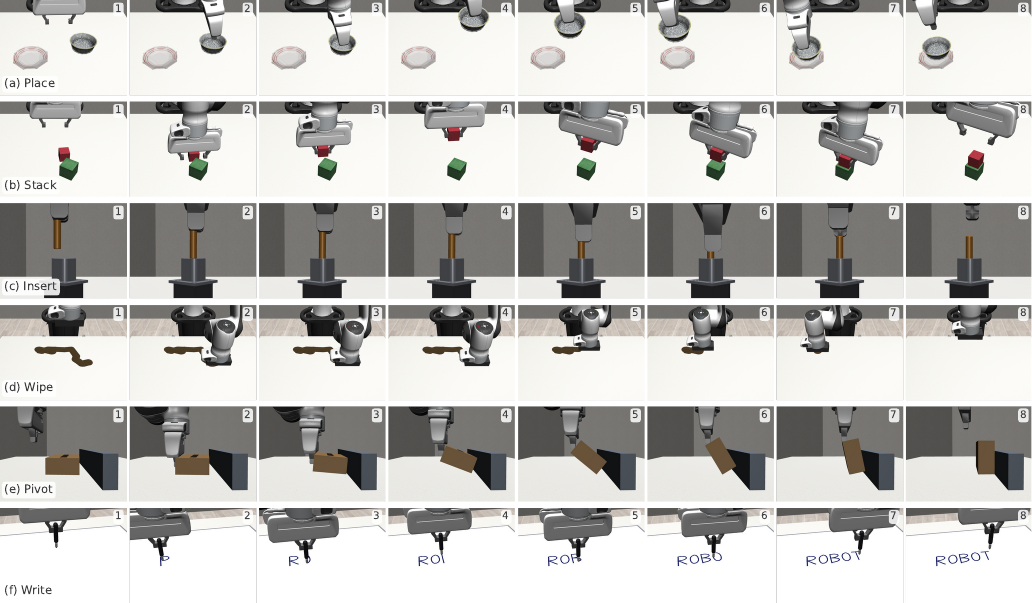}
\caption{Execution and results of Contact Harness (GPT-6 Astra) on the six simulated tasks. (a) Place, (b) Stack, (c) Insert, (d) Wipe, (e) Pivot, and (f) Write each show one run in temporal order from left to right.}
\label{fig:s_sim_results}
\end{figure*}
The benchmark provides 50 human demonstrations per task, 300 in total, with 93195 frames at 10\,Hz, about 2.6 hours. They are collected by teleoperation with a SpaceMouse in the simulated scenes used for evaluation, with random seeds from 100 on, which do not overlap with the evaluation seeds. The robot keeps the Cartesian impedance stiffness used in evaluation, with the damping ratio raised to 2.0 in translation and 4.0 in rotation, and the trained policies are evaluated with the same damping. In Wipe and Write, the tilt of the tool is locked.

The operator saves only successful demonstrations, and stationary frames are removed when saving. In each frame, the observation includes $320 \times 240$ images from the front and wrist cameras and a 13-dimensional state, which covers the end-effector pose, the gripper opening, and the 6-dimensional force and torque. The action includes the target end-effector pose and the gripper command, 7 dimensions in total. The numbers of frames per task are 4972 for Insert, 6429 for Pivot, 9258 for Stack, 10704 for Place, 14018 for Wipe, and 47814 for Write, and the mean demonstration lasts 9.9\,s, 12.9\,s, 18.5\,s, 21.4\,s, 28.0\,s, and 95.6\,s, respectively.

\subsubsection{VLA baselines}
\label{sec:s_res_vla}
We evaluate $\pi_{0.5}$, $\pi_{0.5}$ + force, ForceVLA, and CompliantVLA, trained on the above demonstrations.

\textbf{Training.} For each of the first three methods, we train two models, one on the demonstrations of Place, Stack, Insert, and Pivot and the other on those of Wipe and Write. The models are fine-tuned with LoRA from the pretrained weights for 20000 steps, and the last checkpoint is used. The optimizer is AdamW with a cosine learning rate schedule, 1000 warmup steps, and a peak learning rate of $2.5 \times 10^{-5}$. $\pi_{0.5}$ + force differs from $\pi_{0.5}$ only in that force and torque are concatenated to the proprioceptive state. CompliantVLA adds a module on top of the fine-tuned $\pi_{0.5}$, in which a VLM adjusts the impedance parameters of the robot. The VLM is GPT-6 Astra and updates the impedance every 15 actions, or every 30 actions in Wipe and Write.

\begin{table*}[t]
\centering
\small
\setlength{\tabcolsep}{4pt}
\caption{Per-task results of Contact Harness with different agent models on the six simulated tasks (high reasoning effort, 10 runs per task). Time and the number of decisions are means over successful runs, and force is the median over runs of the largest contact force sustained for at least 0.2\,s.}
\label{tab:s_agents}
\begin{tabular*}{\textwidth}{@{\extracolsep{\fill}}l*{12}{c}@{}}
\toprule
 & \multicolumn{4}{c}{GPT-6 Astra} & \multicolumn{4}{c}{Claude Fable 5.1} & \multicolumn{4}{c}{Claude Fable 5.2} \\
\cmidrule(lr){2-5} \cmidrule(lr){6-9} \cmidrule(lr){10-13}
Task & Succ. & Time (s) & Dec. & Force (N) & Succ. & Time (s) & Dec. & Force (N) & Succ. & Time (s) & Dec. & Force (N) \\
\midrule
Place & 10/10 & 315 & 27.1 & 3.5 & 10/10 & 383 & 12.9 & 4.9 & 10/10 & 337 & 12.6 & 3.8 \\
Stack & 10/10 & 350 & 28.7 & 1.7 & 10/10 & 313 & 12.0 & 2.4 & 10/10 & 326 & 11.7 & 2.3 \\
Insert & 8/10 & 388 & 22.6 & 20.9 & 9/10 & 703 & 21.6 & 20.8 & 9/10 & 676 & 17.7 & 20.9 \\
Wipe & 10/10 & 179 & 10.4 & 9.5 & 10/10 & 362 & 13.2 & 9.3 & 10/10 & 266 & 10.9 & 9.5 \\
Write & 10/10 & 479 & 37.1 & 2.0 & 10/10 & 400 & 13.9 & 2.7 & 10/10 & 451 & 15.3 & 2.7 \\
Pivot & 10/10 & 422 & 42.5 & 6.4 & 10/10 & 416 & 15.2 & 7.0 & 10/10 & 557 & 16.0 & 6.4 \\
\bottomrule
\end{tabular*}
\end{table*}

\textbf{Observations and actions.} Observations consist of the two images and a 7-dimensional state, namely the end-effector pose and the gripper opening. The state of ForceVLA and $\pi_{0.5}$ + force is 13-dimensional and also contains the 6-dimensional force and torque. Actions are absolute end-effector pose targets at 10\,Hz. The model outputs 50 steps at a time, and the first 5 are executed.

\textbf{Evaluation.} Evaluation is capped by the number of action steps, with 350 for Place, 400 for Pivot, 600 for Stack, Insert, and Wipe, and 1800 for Write, corresponding to 35\,s to 180\,s. Except in Write, a rollout ends and counts as successful as soon as the success condition holds. In Write, the rollout runs for the full 1800 steps, and GPT-6 Astra then judges from three views of the whiteboard whether the word ROBOT can be read. The main paper reports seeds 0 to 9 of each task.

\subsection{Results of Contact Harness}
\label{sec:s_res_ours}
\begin{figure*}[p]
\centering
\includegraphics[width=0.92\textwidth]{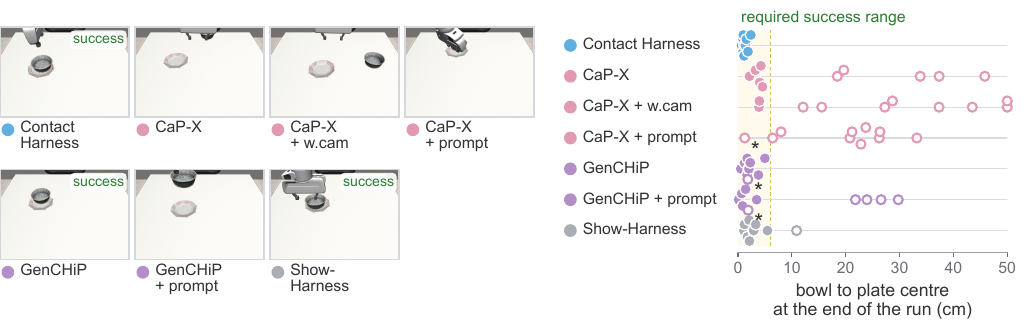}
\caption{Results of each method on Place. The left panel shows the scene at the end of one representative run of each method. In the right panel, each row is a method and each dot is a run, and all methods use the same ten initial states. Filled dots are successful runs and hollow dots are failed runs. The horizontal axis is the horizontal distance between the bowl and the plate center at the end of the run, and the {shaded} region is the range required for success. The three failed runs marked with $\ast$ end within this range. In one, the bowl is still held above the plate (CaP-X + prompt in the left panel). In the other two, it catches on an open finger of the rising gripper (GenCHiP + prompt in the left panel).}
\label{fig:s_base_place}
\vspace{0.3em}
\includegraphics[width=0.92\textwidth]{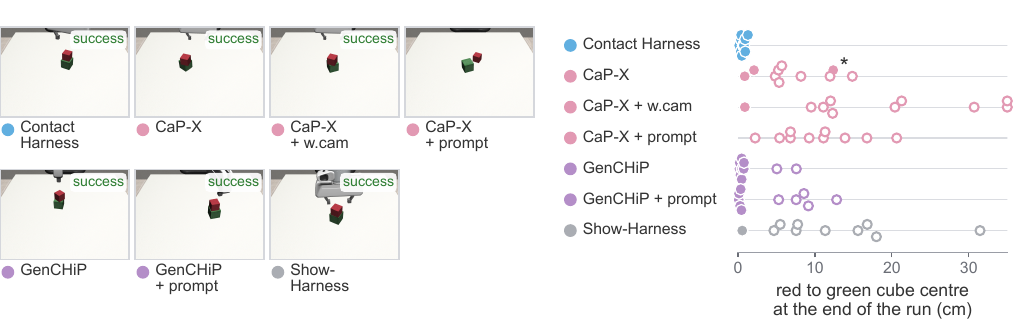}
\caption{Results of each method on Stack. The horizontal axis of the right panel is the horizontal distance between the centers of the red and green cubes at the end of the run. The run marked with $\ast$ was stacked once but moved off later, and it is still marked as a success. The rest is as in \cref{fig:s_base_place}.}
\label{fig:s_base_stack}
\vspace{0.3em}
\includegraphics[width=0.92\textwidth]{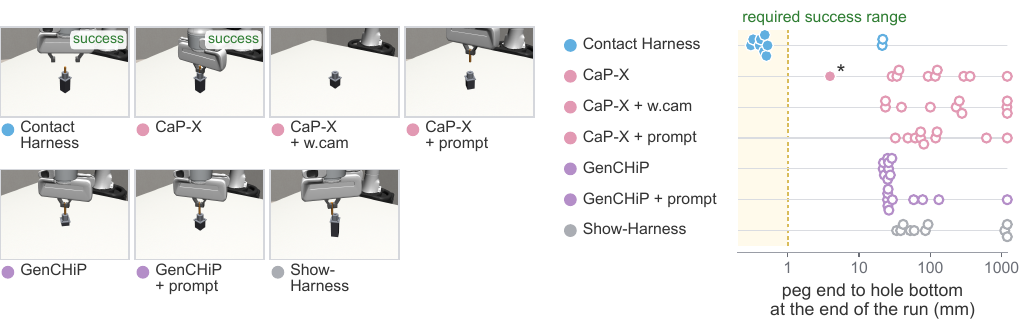}
\caption{Results of each method on Insert. The horizontal axis of the right panel is the distance from the lower end of the peg to the hole bottom at the end of the run, on a logarithmic scale, and the {shaded} region means that the peg is seated at the hole bottom. The run marked with $\ast$ entered the 1\,mm range once but was pulled out later, and it is still marked as a success. The rest is as in \cref{fig:s_base_place}. In the runs beyond 1\,m, the peg drops out of the gripper and ends on the floor beside the table. Points outside the axis range are drawn at the boundary.}
\label{fig:s_base_insert}
\end{figure*}

\begin{figure*}[p]
\centering
\includegraphics[width=0.92\textwidth]{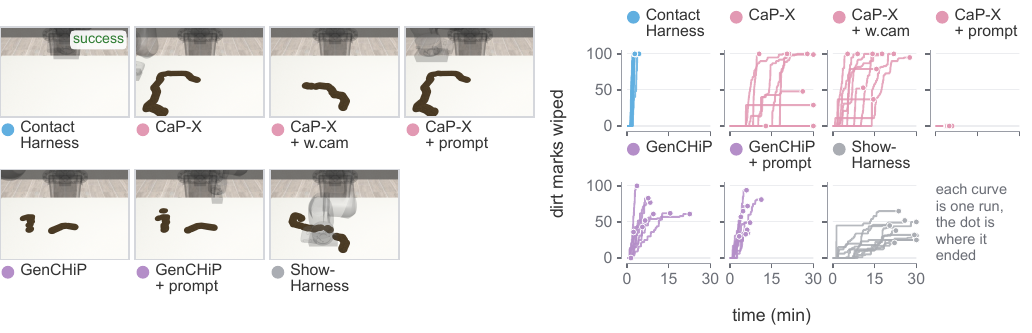}
\caption{Results of each method on Wipe. The table has 100 dirt markers, and success requires removing all of them. The left panel shows the scene at the end of one representative run of each method, where the dark areas are the markers that remain and the robot is drawn semi-transparent. In the right panel, each subplot is a method, each curve is the number of removed markers over time in one run, and the dot marks the end of the run. A curve that reaches 100 has removed all markers. Ten runs are plotted per method, and all methods use the same ten initial states.}
\label{fig:s_base_wipe}
\vspace{0.3em}
\includegraphics[width=0.92\textwidth]{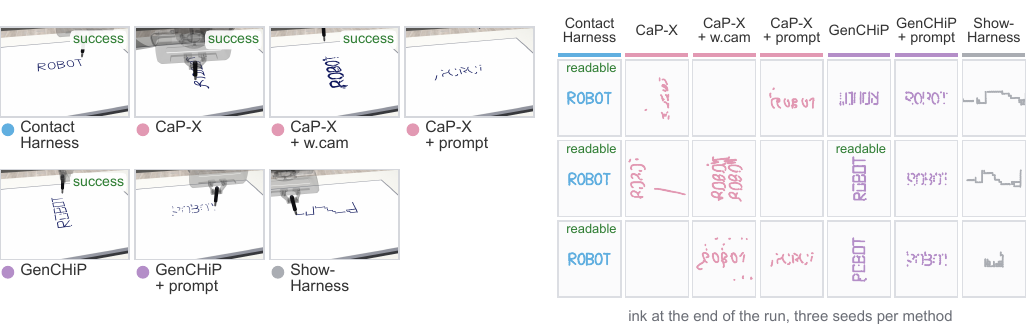}
\caption{Results of each method on Write. The task is to write the word ROBOT on the whiteboard, and a person judges whether the writing is recognizable. The left panel shows the scene at the end of one representative run of each method, with the robot drawn semi-transparent. In the right panel, each column is a method and each row is a random seed, showing all ink at the end of the run. Each cell is 300\,mm wide, and the runs judged recognizable are labeled readable.}
\label{fig:s_base_write}
\vspace{0.3em}
\includegraphics[width=0.92\textwidth]{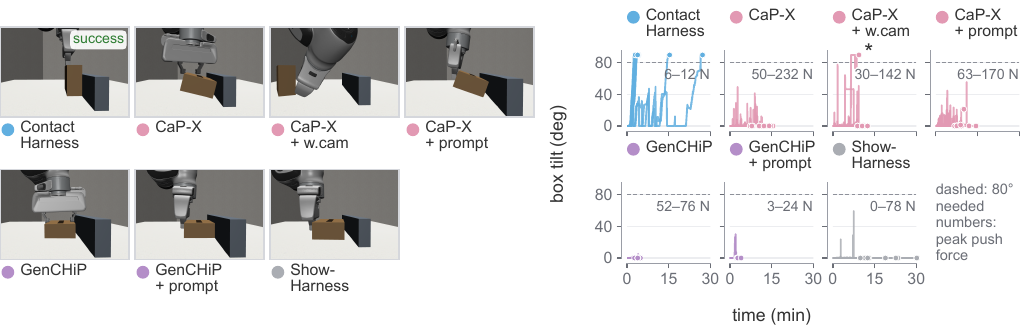}
\caption{Results of each method on Pivot. Success requires the box to stand against the wall and stay still with a tilt angle of at least $80^\circ$. The left panel shows the scene at the moment of maximum tilt in one representative run of each method. In the right panel, each subplot is a method, each curve is the tilt angle of the box over time in one run, the dot marks the end of the run, and the dashed line is the $80^\circ$ threshold. The numbers in each subplot give the range of the peak pushing force over the runs. In the failed run marked with $\ast$, the box stands upright but about 20\,cm away from the wall (CaP-X + w.cam in the left panel). Ten runs are plotted per method, and all methods use the same ten initial states.}
\label{fig:s_base_pivot}
\end{figure*}

\Cref{fig:s_sim_results} shows one run of Contact Harness (GPT-6 Astra) on each of the six simulated tasks, and \cref{fig:s_base_place,fig:s_base_stack,fig:s_base_insert,fig:s_base_wipe,fig:s_base_write,fig:s_base_pivot} show its runs on each task next to those of the baselines. The agent completes all ten runs of every task except Insert, where two runs fail (\cref{fig:s_base_insert}). In two Pivot runs the box falls back several times, and the agent pushes it up again and completes the task (\cref{fig:s_base_pivot}).

\Cref{tab:s_agents} lists the per-task results of Contact Harness with GPT-6 Astra, Claude Fable 5.1, and Claude Fable 5.2, all with high reasoning effort and 10 runs per task, corresponding to the three rows in Table~4 of the main paper. All three models reach 10/10 on Place, Stack, Wipe, Write, and Pivot and differ only on Insert, where Astra reaches 8/10 and both Fable versions reach 9/10. The sustained contact forces are similar across the three models on every task. On Insert, for example, all are about 21\,N, close to the 20\,N soft limit set for the insertion phase. Except on Wipe, Fable 5.1 and 5.2 need fewer decisions than Astra on every task, but each decision takes longer, so the total time is not shorter.

\subsection{Failure analysis of the baselines}
\label{sec:s_fail_base}
This subsection first explains the causes of failure by method and then shows the results of each method by task and analyzes the observed behavior. The per-task numbers behind the analysis are listed in \cref{tab:s_prompt_effect}.

\subsubsection{Limitations of each baseline}
\label{sec:s_fail_methods}

We now discuss the limitations of each baseline on the proposed contact-rich benchmark.

\textbf{Show-Harness.} At every step, the VLM of Show-Harness looks at the images and selects one discrete action, which moves the end effector by 1\,cm along one coordinate axis. Its observations contain no force or tactile information, so it can judge the contact state only from images. Whether an insertion is seated, whether the wiping pad presses on the table, and whether the pen tip touches the board are all hard to see in images, and it often judges the task complete too early.

Most critically, the fixed 1\,cm step is too coarse. It cannot achieve millimeter-level alignment and cannot draw diagonal lines or arcs (\cref{fig:s_base_write}). Shrinking the step much further would in turn require too many agent calls, which lowers efficiency and keeps the task from finishing within the time limit.

\textbf{CaP-X.} In each round, CaP-X generates a code block and executes it open loop without reading force, and it decides the next round from images only after execution ends. Since the environment can be rigid, such open-loop execution can make the peak force shoot above 100\,N (\cref{fig:s_base_pivot}). When the gripper approach is made deliberately slow to avoid overshoot, the repeated code generation for fine-grained adjustment is inefficient and always exhausts the code budget before making meaningful progress.

The wrist camera improves the estimates of marker and tool positions in wiping but does not change the open-loop execution. It therefore helps little in Insert, Write, and Pivot, which need continuous force regulation.

\textbf{GenCHiP.} GenCHiP generates its program once before execution. The agent is not called again during execution and does not see images. The program can end a move on a force or position error condition, but once the move ends, it cannot tell how far the task has progressed. For example, every Insert program contains search motions near the hole opening. When the peg does not drop into the hole during the search, the program is unaware of this. It continues with the next hard-coded push, which naturally fails because the peg is not even aligned, and then ends as usual with the peg still beside the hole opening (\cref{fig:s_base_insert}). Similar problems occur in Wipe, Write, and Pivot, where later motions need to adapt to the state after contact, but all motions are fixed when the program is generated.

Contact Harness avoids these issues. Compliant control and the soft limit constrain the contact force during execution, and before each decision the agent reads images, force, and tactile signals to choose the next action.

\subsubsection{Failure behavior by task}
\label{sec:s_fail_tasks}
This subsection analyzes the behavior of CaP-X, GenCHiP, and Show-Harness on each task. In \cref{fig:s_base_place,fig:s_base_stack,fig:s_base_insert,fig:s_base_wipe,fig:s_base_write,fig:s_base_pivot}, the left panel shows the scene of one representative run of each method, and the right panel plots the results of ten runs of each method, with Contact Harness as the reference. The analysis does not include trained policies.

\begin{table*}[t]
\centering
\small
\setlength{\tabcolsep}{4pt}
\caption{Per-task results of the VLM-based baselines on the six simulated tasks, with 10 runs per task. Time and the number of decisions are means over successful runs and are shown as -- when there is no successful run. Force is the median over runs of the sustained peak contact force.}
\label{tab:s_prompt_effect}
\begin{tabular*}{\textwidth}{@{\extracolsep{\fill}}l*{12}{c}@{}}
\toprule
 & \multicolumn{4}{c}{CaP-X} & \multicolumn{4}{c}{CaP-X + w.cam} & \multicolumn{4}{c}{CaP-X + our prompt skill} \\
\cmidrule(lr){2-5} \cmidrule(lr){6-9} \cmidrule(lr){10-13}
Task & Succ. & Time (s) & Dec. & Force (N) & Succ. & Time (s) & Dec. & Force (N) & Succ. & Time (s) & Dec. & Force (N) \\
\midrule
Place & 5/10 & 375 & 6.4 & 58.0 & 2/10 & 339 & 11.5 & 177.0 & 0/10 & -- & -- & 22.5 \\
Stack & 3/10 & 722 & 8.0 & 48.7 & 1/10 & 274 & 8.0 & 400.0 & 0/10 & -- & -- & 45.6 \\
Insert & 1/10 & 380 & 12.0 & 56.0 & 0/10 & -- & -- & 51.9 & 0/10 & -- & -- & 11.9 \\
Wipe & 4/10 & 1199 & 7.2 & 28.3 & 6/10 & 855 & 4.3 & 31.1 & 0/10 & -- & -- & 0.0 \\
Write & 1/10 & 205 & 6.0 & 10.0 & 1/10 & 364 & 6.0 & 16.1 & 0/10 & -- & -- & 7.0 \\
Pivot & 0/10 & -- & -- & 106.0 & 0/10 & -- & -- & 146.8 & 0/10 & -- & -- & 83.1 \\
\bottomrule
\end{tabular*}

\vspace{0.6em}
\begin{tabular*}{\textwidth}{@{\extracolsep{\fill}}l*{12}{c}@{}}
\toprule
 & \multicolumn{4}{c}{GenCHiP (goal only)} & \multicolumn{4}{c}{GenCHiP + our prompt skill} & \multicolumn{4}{c}{Show-Harness} \\
\cmidrule(lr){2-5} \cmidrule(lr){6-9} \cmidrule(lr){10-13}
Task & Succ. & Time (s) & Dec. & Force (N) & Succ. & Time (s) & Dec. & Force (N) & Succ. & Time (s) & Dec. & Force (N) \\
\midrule
Place & 9/10 & 80 & 1.0 & 13.4 & 5/10 & 100 & 1.0 & 5.9 & 9/10 & 714 & 50.7 & 15.6 \\
Stack & 8/10 & 64 & 1.0 & 7.6 & 5/10 & 87 & 1.0 & 7.4 & 1/10 & 1098 & 64.0 & 13.9 \\
Insert & 0/10 & -- & -- & 4.6 & 0/10 & -- & -- & 1.5 & 0/10 & -- & -- & 21.2 \\
Wipe & 1/10 & 211 & 1.0 & 17.5 & 0/10 & -- & -- & 14.9 & 0/10 & -- & -- & 3.0 \\
Write & 2/10 & 1258 & 1.0 & 14.6 & 0/10 & -- & -- & 10.2 & 0/10 & -- & -- & 2.8 \\
Pivot & 0/10 & -- & -- & 8.4 & 0/10 & -- & -- & 4.0 & 0/10 & -- & -- & 1.1 \\
\bottomrule
\end{tabular*}
\end{table*}
\textbf{Place.} The results on Place are shown in \cref{fig:s_base_place}, with the numbers in \cref{tab:s_prompt_effect}. GenCHiP and Show-Harness both reach 9/10 and CaP-X reaches 5/10. CaP-X does grasp and lift the bowl, and the failure comes from its own check. After grasping, CaP-X localizes the bowl again in the camera view to confirm the grasp. The lifted bowl is occluded by the gripper, and the plate is then often mistaken for the bowl. The program concludes that the bowl is still on the table and opens the gripper to grasp again, and the bowl drops from mid-air. GenCHiP uses the object position directly, and in Show-Harness the model decides every step from images, so neither has this step. CaP-X thus fails on a task without contact. However, the CaP-X paper also reports success rates of only 0.12 to 0.26 under position and task perturbations on LIBERO-PRO~\cite{fu2026capx}.

The wrist camera makes CaP-X worse on this task, at 2/10. CaP-X plans its grasp on the point cloud of the object. With one camera, the cloud mainly contains the side of the bowl facing the camera, and 92\% of the planned grasps lie inside the rim, most of them on that side. The wrist camera looks down on the bowl and adds the whole rim to the cloud. The planned grasps then spread around the bowl, 26\% of them lie outside the rim, and 16\% are less than 1\,cm above the table. For these grasps, the open-loop motion drives the gripper into the table and knocks the bowl away. The median peak force rises from 58\,N to 177\,N, and the bowl is lifted in only five of the ten runs, compared with nine with one camera.

\textbf{Stack.} The results on Stack are shown in \cref{fig:s_base_stack}, with the numbers in \cref{tab:s_prompt_effect}. For CaP-X, only 14 of the 259 grasp attempts in the ten runs pick up the cube. The grasp pose is planned on the point cloud of the cube and executed open loop. In the failed attempts, the gripper comes down beside the cube center, stops above the cube, or closes early in the air. When the cube is lifted in a failed run, it is set down 4 to 6\,cm from the center of the green cube and lands beside it.

Show-Harness spends most of its steps moving back and forth. After a 1\,cm step, it often judges from the images that the held cube is now on the other side of the target and reverses, so only 35\% of its moves in the failed runs contribute to net displacement. Each step also takes one agent call of about 12\,s, and all nine failed runs reach the time limit with the cube not stacked. The released prompt of Show-Harness forbids such back-and-forth motion, which is reasonable for a free-space task like Stack. We removed this rule to allow adjustment under contact, and its side effect appears here.

\textbf{Insert.} The results on Insert are shown in \cref{fig:s_base_insert}, with the numbers in \cref{tab:s_prompt_effect}. For GenCHiP, the peg rests on the top face of the block beside the hole in seven of the ten runs, and in the other three it enters the hole and stops about 3\,mm deep, where the press-fit resistance begins. Its program ends the downward press on a force threshold. Without being told the required force, the agent sets this threshold below the 15\,N of the press fit. In a follow-up experiment not included in the main paper, the task prompt additionally tells the agent that a large force is needed in the push phase. One of ten runs then succeeds by overcoming the press-fit resistance, and the other nine fail to align the peg during the search phase.

Only one of the ten CaP-X runs reaches the hole bottom. The generated code presses down toward position targets without reading force and keeps pressing when the peg is offset from the opening. The contact force exceeds two hundred newtons at its highest, and the peg is pushed aside or slips out of the hand. With Show-Harness, the peg mostly stops 3 to 9\,cm from the hole. Show-Harness judges the relative position of the peg and the hole from images only, and the 1\,cm step cannot achieve millimeter-level alignment.

\textbf{Wipe.} The results on Wipe are shown in \cref{fig:s_base_wipe}, with the numbers in \cref{tab:s_prompt_effect}. A CaP-X run mostly either removes everything or misses entirely. Four of the ten runs remove all markers, three remove none, and the other three remove 29 to 99. Its code gives the whole wiping trajectory at once and executes it by position without adjusting to contact. With the wrist camera, CaP-X removes all markers in six of ten runs.

GenCHiP removes all markers once and 40\% to 80\% in most runs. As its model does not see images, the wiping path is generated from the marker positions and extent given by the interface, and the parts that are missed go unnoticed by both the agent and its code. Show-Harness removes 21\% to 65\%. It moves along the markers step by step, with 1\,cm and one agent call per step, and has covered only part of the markers when the run ends. In some runs it judges the task complete while markers remain. The progress of every run of each method is plotted in the right panel of \cref{fig:s_base_wipe}.

\textbf{Write.} The results on Write are shown in \cref{fig:s_base_write}, with the numbers in \cref{tab:s_prompt_effect}. The actions of Show-Harness are only 1\,cm moves along the coordinate axes, which cannot produce diagonal lines or arcs. This is clearly visible in the left panel of \cref{fig:s_base_write}. The first stroke of R comes out as expected, and Show-Harness even manages to draw the curved part with horizontal and vertical segments. From the diagonal stroke of R on, however, the strokes are misplaced, and what follows no longer resembles any letter. In the conversation record, Show-Harness still judges each letter complete and moves on, which shows that it has lost control of the drawing.

The strokes of CaP-X are mostly scattered and overlapping, and ROBOT is recognizable in only one of ten runs. CaP-X lowers the pen to position targets, and the depth is not adjusted through force control. In five runs, the pen tip is marked as damaged by excessive force (a record only, not a success criterion). The strokes of GenCHiP are more complete in shape. Two runs are judged recognizable, and in the others the strokes break into dotted lines. GenCHiP has two advantages over CaP-X here. Its interface provides the poses of the board and the pen tip, whereas CaP-X has to estimate them from camera images. By the design of the method, its moves can also end on a force condition and comply along the contact direction, so it finds the board by contact and keeps a light pressure while drawing.

\begin{table*}[t]
\centering
\small
\setlength{\tabcolsep}{4pt}
\caption{Per-task ablation results on the four contact tasks (GPT-6 Astra, 10 runs per task). Time and the number of decisions are means over successful runs, force is the median over runs of the sustained peak contact force, and time and decisions are shown as -- when there is no successful run.}
\label{tab:s_ablation}
\begin{tabular*}{\textwidth}{@{\extracolsep{\fill}}l*{8}{c}@{}}
\toprule
 & \multicolumn{4}{c}{Full method} & \multicolumn{4}{c}{w/o force observations} \\
\cmidrule(lr){2-5} \cmidrule(lr){6-9}
Task & Succ. & Time (s) & Dec. & Force (N) & Succ. & Time (s) & Dec. & Force (N) \\
\midrule
Insert & 8/10 & 388 & 22.6 & 20.9 & 7/10 & 491 & 28.6 & 14.8 \\
Wipe & 10/10 & 179 & 10.4 & 9.5 & 7/10 & 194 & 9.1 & 9.3 \\
Write & 10/10 & 479 & 37.1 & 2.0 & 9/10 & 446 & 36.0 & 2.8 \\
Pivot & 10/10 & 422 & 42.5 & 6.4 & 10/10 & 338 & 32.1 & 7.3 \\
\bottomrule
\end{tabular*}

\vspace{0.6em}
\begin{tabular*}{\textwidth}{@{\extracolsep{\fill}}l*{8}{c}@{}}
\toprule
 & \multicolumn{4}{c}{w/o skills} & \multicolumn{4}{c}{w/o compliance \& soft limit} \\
\cmidrule(lr){2-5} \cmidrule(lr){6-9}
Task & Succ. & Time (s) & Dec. & Force (N) & Succ. & Time (s) & Dec. & Force (N) \\
\midrule
Insert & 0/10 & -- & -- & 7.0 & 0/10 & -- & -- & 86.6 \\
Wipe & 0/10 & -- & -- & 5.1 & 9/10 & 188 & 12.6 & 12.1 \\
Write & 9/10 & 287 & 12.8 & 2.3 & 10/10 & 356 & 34.1 & 5.3 \\
Pivot & 8/10 & 155 & 11.8 & 8.0 & 6/10 & 294 & 36.5 & 10.2 \\
\bottomrule
\end{tabular*}
\end{table*}

\textbf{Pivot.} The results on Pivot are shown in \cref{fig:s_base_pivot}, with the numbers in \cref{tab:s_prompt_effect}. CaP-X can lift the box but cannot push it all the way upright. In most runs it lifts one side of the box, but the maximum tilt varies widely across seeds, from near 0 to $50^\circ$, and never reaches $80^\circ$, after which the box falls back onto the table. During pivoting, the magnitude and direction of the pushing force must change with the tilt of the box, whereas each CaP-X code block executes preset position targets. The instantaneous peak pushing force is mostly above one hundred newtons, far above the 14\,N bound. CaP-X with the wrist camera stands the box up once, but the box is about 24\,cm from the wall and does not meet the condition of standing against the wall.

With GenCHiP the box is hardly lifted. Its programs all choose to grasp the box, lift it, and rotate it upright instead of pushing against the wall, and the model does not see images and cannot notice that the box was not lifted. With the task prompt, the GenCHiP program switches to closing the fingers and pushing against the wall, following the rising edge of the box along a precomputed path. In two runs the box is lifted to about $30^\circ$ but not pushed further upright, because the precomputed path cannot adapt to the actual pose of the box. With Show-Harness the box is not lifted in most runs. It judges the task complete while the box still lies flat, and in two runs it lifts the box to $25^\circ$ and $60^\circ$ before the box falls back.

\subsubsection{Effect of the task prompt on CaP-X and GenCHiP}
\label{sec:s_fail_prompt}
The task prompt is written for the agent of Contact Harness. Before giving it to the baselines, we adapt it in three ways. First, we keep the task goal, the initial state, the step-by-step strategy, and the completion condition. Second, we remove what only Contact Harness can use, namely its command and observation interface and the force and tactile thresholds. Third, for GenCHiP, we also remove the checks that rely on images, which it cannot execute.

In Table~4 of the main paper, adding the task prompt does not improve the success rate of CaP-X or GenCHiP, and CaP-X is at 0/10 on all six tasks (\cref{tab:s_prompt_effect}). Again, the task prompt is written for the agent of Contact Harness. This agent obtains observations right after every step, can confirm contact with force and tactile readings, and is protected by the soft limit. An extra confirmation therefore costs it almost nothing, and the prompt explicitly asks it to confirm before continuing in several steps. Neither of the two methods has these conditions, and the requirement to confirm before continuing leaves them stuck at the checks.

\textbf{CaP-X.} CaP-X generates at most 12 code blocks per attempt. Each confirmation uses up one block, and it can only confirm from images, which often gives no conclusion. With the task prompt, the proportion of code blocks whose comments contain words such as `confirm' or `check' rises from 10\% to 32\%. Wipe is the clearest case. The prompt asks to confirm from images that the wiping pad touches the table before sliding, which is hard to tell from the third-view image. As a result, all ten runs get stuck in the checking phase and use up the 12 code blocks without removing a single marker (\cref{fig:s_base_wipe}). Some other tasks make progress but stop before success. In Insert, for example, four runs align the peg to within 1\,mm of the hole axis, after which the code budget runs out.

\textbf{GenCHiP.} A GenCHiP program is generated once and executed open loop. With the task prompt, the model writes the requirement to confirm before continuing as assertions, and the program aborts when an assertion fails. Of the ten runs per task, 6 abort in Place, 5 in Stack, 7 in Insert, 4 in Write, 1 in Pivot, and none in Wipe. In Place and Stack, four of the five failed runs abort at an assertion around the grasp. In Insert, the aborts mostly occur at the assertion that the peg tip fills the hole opening, before the press is executed. Pivot is the exception. The program switches to closing the fingers and pushing against the wall, and two runs lift the box to about $30^\circ$ (\cref{fig:s_base_pivot}).

\textbf{Show-Harness.} Show-Harness has no variant with our prompt skill. For the four contact tasks, we write only a one-paragraph prompt with the manipulation goal and strategy into its task sentence and do not give it the full task prompt. Its control prompt is also rewritten. However, we observe that its failures mainly come from the coarse 1\,cm step and the lack of force or tactile information in its observations, which a prompt cannot change.

\textbf{Contact force.} In Table~4 of the main paper, the contact force of CaP-X and GenCHiP is lower with the task prompt than with only the task goal, at 28.4\,N versus 51.2\,N for CaP-X and 7.3\,N versus 11.0\,N for GenCHiP. CaP-X cannot read force, and the force is lower because there is less contact. In Wipe, for example, the CaP-X wiping pad hardly touches the table and the median contact force is 0.

\subsection{Ablation}
\label{sec:s_res_abl}
\Cref{tab:s_ablation} lists the per-task results of the four configurations in Table~5 of the main paper, with 10 runs per task for each configuration, judged with the force conditions of \cref{sec:s_eval_sim}.

Without force observations provided to the agent, Insert reaches 7/10. As stated in the main paper, these insertions succeed in a different phase. With the full method, seven of the eight successful runs insert the peg during the alignment phase, where the agent is supposed to use the wrench readings. Without force observations, none of the seven successful runs completes in this phase, and all rely on the final push specified in the task prompt. In Wipe, three runs do not press on the table and remove only 11 to 32 markers with a force of at least 5\,N. In Write, one run is unreadable, and Pivot is unaffected. These results suggest that, even under compliant execution, visual evidence alone is not enough for force-guided corrections, unless the force can be clearly inferred from the object state, such as the tilt angle of the box in Pivot.

Without the textual skills, none of the ten Insert runs reaches the hole bottom. In Wipe, all ten runs remove all 100 markers and count as successful under the criterion of the method comparison. They fail only the additional force condition, because the recorded wiping force stays at 2 to 4\,N and at most 7 markers are removed with at least 5\,N. Without the skill, the agent is not told that wiping needs pressure, and it follows what the simulator shows, where a marker disappears on light contact.

Without compliant execution and the soft limit, all ten Insert runs fail. In eight of them, the sustained peak contact force is between 76\,N and 103\,N, above the 50\,N bound, and the other two do not reach the hole bottom. In Pivot, four runs have a pushing force between 15\,N and 18\,N, above the 14\,N bound. In Wipe, one run does not meet the pressing force condition, and Write is unaffected. These observations suggest that, even when the agent knows the contact force and how to regulate it, its response is too slow to prevent excessive contact force in a rigid environment.

To this end, the simulation ablations show that a high success rate on Insert requires the agent to know the contact force, to know how to adjust it, and to act through a compliant interface that regulates the force at high frequency.

\begin{table*}[t]
\centering
\footnotesize
\setlength{\tabcolsep}{4pt}
\renewcommand{\arraystretch}{1.15}
\caption{The complete course of one successful run of simulated Insert (GPT-6 Astra).}
\label{tab:s_example_sim}
\begin{tabularx}{\textwidth}{c>{\raggedright\arraybackslash}X>{\raggedright\arraybackslash}p{5.6cm}}
\toprule
Step & Agent message (verbatim) & Commands issued next \\
\midrule
1 & I'll read the two project prompts, then use their instructions to guide the peg insertion. & Reads the system prompt and the task prompt, requests the images of the three cameras and the force and tactile readings, and zeroes the force and tactile readings \\
2 & The peg is clear of the block, and both sensors are calibrated. The block has a small top face; I'm moving closer with a slight offset from the hole before establishing surface contact. & 3 displacement commands that approach and touch the top face of the block step by step, with images and readings requested after each command \\
3 & Both wrist views and the head view confirm the peg tip is resting on the block's top rim, with the fingers clear. I've recorded that height. A gap remains beside the peg, so the entry gate has not passed; I'm continuing the surface search at 10 N. & 1 lateral displacement command of 5\,mm \\
4 & The fresh zoomed entry check passes in both wrist views: the peg fills the opening with no visible gap. I'm raising the soft limit to 20 N and starting axial insertion, checking all three views after each adjustment. & Sets the soft limit to 20\,N and issues 3 displacement commands along the hole axis, with a trial rotation of $3^\circ$ that is then reversed \\
5 & The peg remains engaged in all three views, but it has not seated deeply. The tilt trial increased tactile loading, so I reversed it. I'm finishing the permitted alignment adjustments before the prescribed release-and-push step. & 2 small alignment displacement commands, then retreats, opens the gripper, and retreats again \\
6 & The peg stayed in the hole after release. The measured insertion travel was about 25 mm, and a section still protrudes. The cable-free peg is centered beneath the cleared gripper; I'll close the fingers and perform the specified 7 cm axial push. & Closes the fingers, pushes down along the hole axis under the 20\,N soft limit, retreats and opens the gripper, and repeats once \\
7 & The second push reached the same seating height under about 20 N, with lateral force below 1 N. The peg remains upright and engaged. I'm removing the load and checking that it stays seated. & Restores the soft limit to 10\,N, retreats, and opens the gripper \\
8 & The peg is inserted and remains stable after release. I withdrew the gripper and restored the soft force limit to 10 N. & Run ends \\
\bottomrule
\end{tabularx}
\end{table*}
\begin{table*}[t]
\centering
\footnotesize
\setlength{\tabcolsep}{4pt}
\renewcommand{\arraystretch}{1.15}
\caption{The complete course of one successful run of real-robot USB-A insertion (Rokae, GPT-6 Astra).}
\label{tab:s_example_real}
\begin{tabularx}{\textwidth}{c>{\raggedright\arraybackslash}X>{\raggedright\arraybackslash}p{5.6cm}}
\toprule
Step & Agent message (verbatim) & Commands issued next \\
\midrule
1 & I'll read the two requested instruction files, pay close attention to their important notes, and use them to insert the USB-A cable into the left socket. I won't use any skills. & Reads the two prompts, requests observations, and zeroes the force and tactile readings \\
2 & The connector is USB-A, and the target is the left of the two larger USB sockets. The cable is clear of the hub, and both force and tactile baselines are calibrated. I'm raising the connector and approaching while checking the side view for height. & 5 displacement commands that approach the socket in free space \\
3 & The tip is contacting the hub face above the left USB-A opening; both wrist views and the increased force support that. I've recorded this surface position and am sliding downward toward the opening under the 10 N limit. & 5 displacement commands that slide along the housing surface toward the opening \\
4 & The connector has entered the left socket: the exposed metal has shortened, and the measured forward travel exceeds 5 mm from the recorded surface position. I'm switching to the specified 20 N limit and checking small alignment corrections before release. & Sets the soft limit to 20\,N and issues 5 displacement commands for small alignment adjustments \\
5 & The connector remains engaged, and lateral force has dropped to about 1 N. I'm finishing the limited alignment checks, then I'll follow the required withdrawal, release, and cable regrasp sequence for the final push. & 4 displacement commands and 1 gripper command that finish the alignment, then retreat and release \\
6 & The USB-A cable is in the left socket and stayed in place after release. Stopping here. & Run ends \\
\bottomrule
\end{tabularx}
\end{table*}
\section{Complete Run Examples}
\label{sec:s_example}
This section gives the records of two complete runs. \Cref{tab:s_example_sim} is a run of simulated Insert that takes 372\,s with 20 motion command calls. \Cref{tab:s_example_real} is a run of real-robot USB-A insertion that takes about 330\,s from reading the prompts to confirming the insertion, and it is one of the 10 trials analyzed in \cref{sec:s_real_evidence}. In both tables, the middle column is the message the agent writes during the run, quoted verbatim, and the right column summarizes the commands it issues next.

\section{Token Usage}
\label{sec:s_token}
This section reports the token usage of Contact Harness on the real robot (\cref{sec:s_token_real}) and of all VLM-based methods in simulation (\cref{sec:s_res_token}).

\subsection{Real robot}
\label{sec:s_token_real}

The token usage on the real robot is reported for USB-A insertion on the Rokae platform. It is taken from the session records of the 10 trials in \cref{fig:evidence} and is counted in the same way as in simulation (\cref{sec:s_res_token}). A trial uses about 1.62 million tokens on average, with a minimum of 1.01 million and a maximum of 2.71 million, and 95\% of the input is cached. The agent makes 20.4 decisions per trial on average, and trials with more decisions use more tokens. The two trials with 16 decisions use 1.01 and 1.14 million tokens, and the three trials with 26 or 27 decisions use 2.50 to 2.71 million. This usage is close to the 1.49 million tokens per run of Contact Harness (Astra) in simulation (\cref{tab:s_tokens}).

\subsection{Simulation}
\label{sec:s_res_token}
\Cref{tab:s_tokens} reports the token usage of each method in simulation. The token count is the sum of the input and output of all agent calls in a run, with cached input included. Usage per run is averaged within each task and then over the six tasks. Usage per success is the total number of tokens over all runs divided by the number of successes. Most VLA baselines do not call a VLM and are not in the table.

Contact Harness uses about 1.5 to 1.9 million tokens per run, and 96\% of its input is cached. It requests images and readings many times in a run, and every decision carries the whole preceding conversation, so it uses more input tokens than the other methods. CaP-X uses about 0.2 million per run and Show-Harness about 0.62 million. GenCHiP calls the model once and uses about 10 thousand. Per success, Show-Harness uses about 3.74 million, more than the 1.55 to 1.93 million of Contact Harness, CaP-X about 0.82 to 1.19 million, and GenCHiP about 30 to 60 thousand.

\par\medskip
{\centering
\begin{minipage}{\colw}
\small
\setlength{\tabcolsep}{3pt}
\captionof{table}{Token usage of each method in simulation, in millions. Cached is the share of input tokens served from the cache.}
\label{tab:s_tokens}
\begin{tabular*}{\linewidth}{@{\extracolsep{\fill}}lccc@{}}
\toprule
Method & Per run & Cached & Per success \\
\midrule
CaP-X & 0.19 & 47\% & 0.82 \\
CaP-X + w.cam & 0.20 & 46\% & 1.19 \\
CaP-X + our prompt skill & 0.21 & 47\% & -- \\
GenCHiP (goal only) & 0.009 & 56\% & 0.027 \\
GenCHiP + our prompt skill & 0.010 & 53\% & 0.059 \\
Show-Harness & 0.62 & 61\% & 3.74 \\
Contact Harness (Astra) & 1.49 & 96\% & 1.55 \\
Contact Harness (Fable 5.1) & 1.90 & 96\% & 1.93 \\
Contact Harness (Fable 5.2) & 1.83 & 96\% & 1.86 \\
\bottomrule
\end{tabular*}
\end{minipage}\par}\medskip

\section{Limitations}
\label{sec:s_limit}
The main limitation of Contact Harness is its long execution time. In Table~1 of the main paper, a connector insertion takes 378\,s on average, of which about 90\% is spent on model inference and obtaining observations. With the fast mode and memory, the average time for IEC C13 drops to 96\,s, which is still slower than policies that output actions directly~\cite{arachchige2025sail,guo2025demospeedup,wu2026speedup,jing2026tempovla,hu2026freespeed}.

Deployment in more complex real-world scenes faces further challenges. First, the system depends on a remote model service, and service interruptions or stalls put the execution on hold and sometimes cause a failure directly. Second, perception is limited by the cameras~\cite{hu2026waveman}. Small plugs such as Micro USB occupy very few pixels in the image, and the plug and the socket may also be occluded, which makes it hard for the agent to judge the alignment. Third, token usage is high, with a real-robot USB-A insertion using about 1.6 million tokens on average.

Fourth, some sockets are still hard to handle even with prompt engineering. A socket of this kind, such as the HDMI socket and connector pair we have at hand, has a hidden ring, a second metal edge that lies deep inside the opening behind the visible contact surface and cannot be seen from any camera view. When the plug is halfway in and its advancing path is not exactly along the center of the socket, it catches on this edge. The contact force then rises to the soft limit and the insertion advances no further, exactly as when the plug is fully inserted, so the agent cannot tell the two cases apart from the force. Vision can separate them, for example when metal parts of the plug are still partially exposed outside the socket, but not when the large fingertips block the camera views. A promising solution is a programmed skill that regulates the force in code under light contact and vibrates the plug laterally to escape the second edge before advancing further. Agent systems that call such modular motion skills exist for cable insertion~\cite{shan2026llm}, and haptic estimation with adaptive stiffness at the controller level prevents jamming in tight-tolerance tool use~\cite{yang2026adaptive}. This requires a high control frequency with controller-level adaptation and is therefore hard to achieve with a purely agentic harness.

Fifth, the current surface search method under contact does not work for connectors with an uneven contact surface, because the sliding search can get stuck on the protruding parts. The prompt can instead ask the agent to withdraw and align the connector precisely by vision before making contact again, so that the connector fits in directly. This solves the problem but requires connector-specific prompt engineering, so it is not included in the paper.

Contact Harness currently requires an arm with direct wrist force sensing or joint-level torque sensing and compliant control. Sensorless contact estimation removes this requirement. It estimates the external wrench from the internal signals of the robot~\cite{hu2018contact,yilmaz2020neural,shan2024fine,shan2024sensorless}, reaches the accuracy needed for a pin insertion~\cite{shan2024fine,ge2025filic}, and recalibrates within seconds when the payload changes~\cite{shan2024fast}. Combining it with Contact Harness brings the method to low-cost arms without a force sensor or a built-in impedance controller~\cite{oh2026factr2,dou2026neuralactuator}.

\section{Prompts of All Tasks}
\label{sec:s_prompts}
\begin{figure}[htbp]
\centering
\includegraphics[page=1,width=0.96\textwidth]{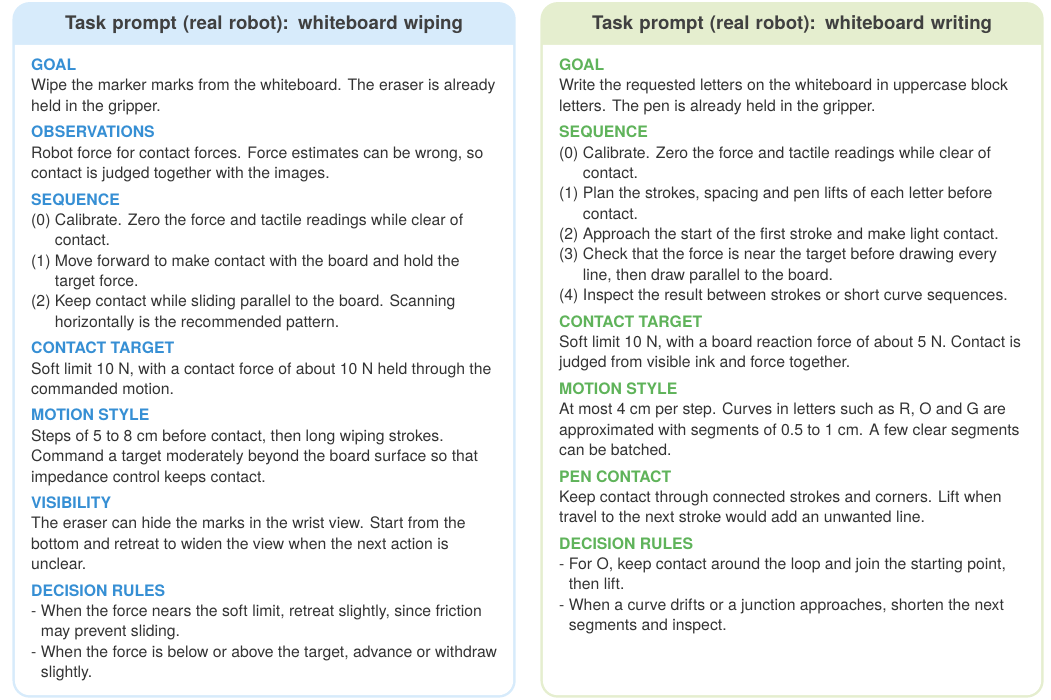}
\caption{Summaries of the task prompts for real-robot whiteboard wiping (left) and writing (right).}
\label{fig:s_prompts_real1}
\end{figure}

This section gives the summaries of the task prompts of the remaining tasks, in the same form as the right panel of \cref{fig:s_prompts} and likewise condensed from the prompts used in the experiments. The connector insertion task prompt is given in \cref{fig:s_prompts}. \Cref{fig:s_prompts_real1,fig:s_prompts_real2} are the prompts of the four real-robot tasks of whiteboard wiping, writing, peeling, and pivoting. \Cref{fig:s_prompts_sim1,fig:s_prompts_sim2,fig:s_prompts_sim3} are the prompts of the six tasks in the simulated benchmark. Real-robot stacking and pick-and-place have no separate task prompt and are specified by a short task description.

Notably, for stacking and pick-and-place on the real Franka, the agent is given no specific pick-and-place task prompt. It is asked to follow the motion style of the insertion task prompt and is told only that the task is pick-and-place instead of insertion. For Place and Stack in simulation, the task prompts are generated by an AI agent that takes the existing prompts of the contact tasks as reference, with no manual modification.

\begin{figure}[htbp]
\centering
\includegraphics[page=2,width=0.96\textwidth]{figs/figureS2/prompt_panels_all.pdf}
\caption{Summaries of the task prompts for real-robot peeling (left) and pivoting (right).}
\label{fig:s_prompts_real2}
\end{figure}
\begin{figure}[htbp]
\centering
\includegraphics[page=3,width=0.96\textwidth]{figs/figureS2/prompt_panels_all.pdf}
\caption{Summaries of the task prompts for simulated Insert (left) and Wipe (right).}
\label{fig:s_prompts_sim1}
\end{figure}
\begin{figure}[htbp]
\centering
\includegraphics[page=4,width=0.96\textwidth]{figs/figureS2/prompt_panels_all.pdf}
\caption{Summaries of the task prompts for simulated Write (left) and Pivot (right).}
\label{fig:s_prompts_sim2}
\end{figure}
\begin{figure}[htbp]
\centering
\includegraphics[page=5,width=0.96\textwidth]{figs/figureS2/prompt_panels_all.pdf}
\caption{Summaries of the task prompts for simulated Place (left) and Stack (right).}
\label{fig:s_prompts_sim3}
\end{figure}
\FloatBarrier

\end{document}